\documentclass[10pt,twocolumn,letterpaper]{article}

\usepackage{cvpr}              
\usepackage{algorithm}
\usepackage[noend]{algpseudocode}
\usepackage{amsmath,amssymb}
\usepackage{booktabs}
\usepackage[table]{xcolor} 
\usepackage{float}
\usepackage{tabularx}
\definecolor{VLABlue}{RGB}{241,246,250}
\definecolor{WAMGreen}{RGB}{241,248,243}
\definecolor{AblYellow}{RGB}{250,247,232}
\definecolor{DeepRed}{RGB}{140,45,45}
\usepackage{multirow}
\usepackage{graphicx}
\usepackage{cuted}
\usepackage{caption}
\usepackage{pifont}
\usepackage{dblfloatfix}
\definecolor{ObservationBlue}{HTML}{0078B0}
\definecolor{GaussTwinPurple}{HTML}{8342B8}
\definecolor{SimulationYellow}{HTML}{967000}
\definecolor{FailureRed}{HTML}{C73532}
\definecolor{SuccessGreen}{HTML}{348019}

\usepackage{cuted} 

\usepackage{currfile} 

\usepackage{caption} 

\definecolor{cvprblue}{rgb}{0.21,0.49,0.74}
\usepackage[pagebackref,breaklinks,colorlinks,allcolors=cvprblue]{hyperref}

\def\confName{CVPR}
\def\confYear{2027}

\title{SpatialHarness: Augmenting Spatial Observability for Precise \\Robotic Manipulation}

\author{
\textbf{Jiayu Wang}$^{1,2,*}$ \hspace{0.5em}
\textbf{Yue Yu}$^{1,2,*}$ \hspace{0.5em}
\textbf{Bin Zhu}$^{3}$ \hspace{0.5em}
\textbf{Zhiyao Yang}$^{1}$ \hspace{0.5em}
\textbf{Jingjing Chen}$^{1,2,\dagger}$ \hspace{0.5em}
\textbf{Yu-Gang Jiang}$^{1,2}$ \\[0.4em]
\mbox{$^{1}$Fudan University \quad
$^{2}$Shanghai Key Laboratory of Multimodal Embodied AI} \\
$^{3}$Singapore Management University \\[0.2em]
{\tt\small \{jiayuwang25,yuy24\}@m.fudan.edu.cn} \\
{\tt\small binzhu@smu.edu.sg \quad
\{chenjingjing,ygj\}@fudan.edu.cn}
}

\begin{document}
\maketitle
\begingroup
\makeatletter
\renewcommand{\thefootnote}{}
\renewcommand{\@makefntext}[1]{\parindent=0pt\noindent #1}
\footnotetext[0]{%
$^{*}$ Equal contribution. \\
$^{\dagger}$ Corresponding author.
}
\makeatother
\endgroup

\begin{abstract}
Frontier multimodal foundation models (e.g., GPT-6 Astra) have recently shown strong potential for direct robotic control, yet their performance on precise manipulation remains limited. We argue that an important source of failure is not necessarily insufficient policy capability, but insufficient spatial observability, where task-critical spatial relationships may be poorly revealed by the existing physical camera setup. We introduce SpatialHarness, a test-time embodied harness that augments the spatial observability of frozen multimodal policies without policy fine-tuning or additional physical sensors. SpatialHarness constructs an online synchronized simulated environment and develops runtime spatial inspection interface, providing complementary virtual views and explicit geometric queries to reveal spatial relationships essential for precise manipulation. To maintain consistency between simulation and reality during robotic interaction, we develop an interaction-aware scene synchronization mechanism that combines state propagation, hypothesis-based interaction replay, and observation-grounded correction across static, held, and transition modes. We evaluate SpatialHarness on four real-robot manipulation tasks under both direct and agentic control paradigms. Under direct control, SpatialHarness substantially improves the average success rate of GPT-6 Astra low from 18.3\% to 83.3\%. When integrated into Harness VLA under agentic control with GPT-6.1 Sol high as Action Planner, it further increases the average success rate from 73.3\% to 96.7\%. These results demonstrate that augmenting spatial observability at test time can substantially improve precise robotic manipulation without modifying the underlying policy. Project website: \url{https://emilia113.github.io/SpatialHarness/}.
\end{abstract}

\section{Introduction}

\begin{figure}[t]
  \centering
  \makebox[\linewidth][c]{%
    \includegraphics[width=\linewidth]{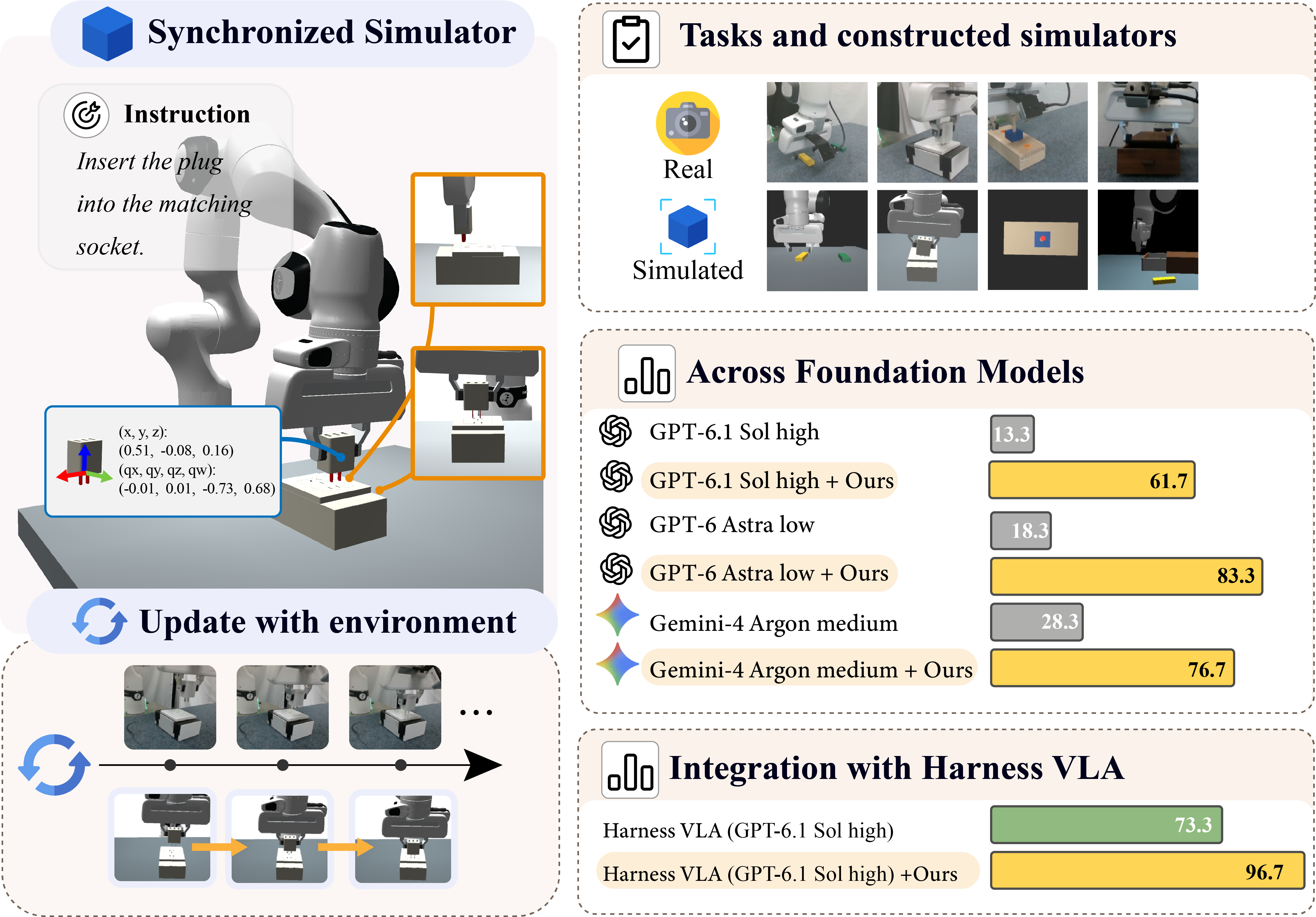}
  }
  \caption{During plug insertion, the simulator tracks the physical scene and provides virtual views and object poses (left). Each real task is paired with its simulated counterpart (top right).
  Bars report mean success across the four tasks. With SpatialHarness (yellow), mean success rises from 18.3\% to 83.3\% for GPT-6 Astra low under direct control and from 73.3\% to 96.7\% for Harness VLA under agentic control using GPT-6.1 Sol high.}
  \label{fig:teaser}
  \vspace{-10pt}
\end{figure}

Recent advances in multimodal foundation models have opened new possibilities for robotic manipulation.  Frontier models such as GPT-6 Astra~\cite{gpt6astra} and Gemini-4 Argon~\cite{kavukcuoglu2026gemini4argon} have demonstrated the ability to directly generate robot actions from visual observations and language instructions, enabling closed-loop manipulation without task-specific policy training. This emerging paradigm offers a promising alternative to conventional robotic foundation models~\cite{pmlr-v229-zitkovich23a,pmlr-v270-kim25c,black2024pi0,Ghosh-RSS-24,intelligence2025pi05}, which typically require additional training on robotic data. 

Two emerging paradigms have enabled frontier multimodal foundation models to perform robotic manipulation: \textit{direct control}, where models generate robot actions directly from visual observations and language instructions~\cite{gpt6astra, kavukcuoglu2026gemini4argon, zhang2026unexpected}, and \textit{agentic control}, where models leverage embodied harnesses with perception tools, action interfaces, and interaction feedback to guide robotic execution~\cite{zhang2026harnessvla, chen2026show}.  Despite their promising capabilities, both paradigms still struggle with precise manipulation tasks~\cite{zhang2026unexpected, team2026systematically}, such as plug insertion and placement onto narrow structures, which require accurate reasoning about spatial relationships. A key limitation lies in insufficient spatial observability, where task-critical geometric relationships are difficult to infer from physical camera observations due to occlusion, unfavorable viewpoints, or perspective ambiguity. Consequently, manipulation failures may arise not only from limitations in the models' reasoning and control capabilities, but also from inadequate spatial information available during action decision. While existing embodied harnesses primarily improve action interfaces and interaction context, recent real-to-sim and digital-twin approaches~\cite{cai2026gausstwin, huang2026synctwinfastdigitaltwin, kapelyukh2026test} demonstrate the potential of simulated environments for representing and tracking physical scenes. This motivates a fundamental question: \textit{Can we improve precise robotic manipulation by augmenting spatial observability at test time, without retraining the policy or modifying the physical sensing setup?}

To address this question, we introduce \textit{SpatialHarness}, a test-time embodied harness that augments the spatial information available to frozen multimodal policies through a synchronized simulated environment. As illustrated in \Cref{fig:teaser}, SpatialHarness transforms simulation into a runtime spatial inspection interface, providing complementary virtual views and explicit geometric queries to reveal task-critical spatial relationships that are difficult to infer from physical cameras. For example, a virtual top-down view can reveal the alignment between a hole and its target peg, while geometric queries provide precise measurements of distances and relative orientations for insertion and placement tasks. A key technical challenge is maintaining alignment between simulation and reality during robotic execution, where pose estimation errors and imperfect contact modeling can cause substantial state divergence. To address this, we develop an interaction-aware scene synchronization mechanism that combines state propagation with observation-grounded correction across three interaction modes: static, held, and transition. In particular, hypothesis-based replay during grasping and releasing, followed by geometric refinement using real RGB-D observations, helps maintain consistent scene states. By grounding augmented spatial information in physical execution, SpatialHarness enables more precise manipulation under both direct and agentic control paradigms, without policy fine-tuning or modifications to the physical sensing setup.

We evaluate SpatialHarness on four real-robot manipulation tasks spanning local geometric alignment, object-relative placement, and articulated-object interaction, under both direct and agentic control frameworks, as shown in Fig.~\ref{fig:teaser}. Under direct control, SpatialHarness improves the average success rate of the GPT-6 Astra low from 18.3\% to 83.3\%, including a substantial improvement from 0\% to 100\% on Hanoi Tower. Under agentic control, integrating SpatialHarness into Harness VLA further increases the average success rate from 73.3\% to 96.7\%. These improvements are achieved without policy fine-tuning or changes to the physical camera setup, highlighting the importance of spatial observability as a test-time factor in fine robotic manipulation.

Our contributions are as follows:

\begin{itemize}
  \item We introduce SpatialHarness, an embodied harness that enhances precise robotic manipulation by augmenting frozen multimodal policies with task-relevant virtual views and structured geometric information, without policy fine-tuning or additional physical sensors.
  
  \item We develop an online synchronization mechanism combining interaction-dependent state propagation, hypothesis-based replay, and observation-grounded correction to maintain alignment between simulated and physical environments during robotic manipulation.  

  \item We demonstrate substantial improvements in success rate across real-robot precise manipulation tasks, different frontier multimodal models, and both direct and agentic control paradigms, highlighting the effectiveness of test-time spatial observability augmentation.
 
\end{itemize}

\section{Related Work}
\subsection{Frontier Multimodal Models and Embodied Harnesses}
Beyond vision-language-action models~\cite{pmlr-v229-zitkovich23a,pmlr-v270-kim25c,black2024pi0,wang2026unified} and world action models~\cite{li2026causal,ye2026world,kim2026cosmos,bi2026motus,wang2026mowam}, recent frontier multimodal models~\cite{gpt6astra,anthropic2026claudeopus55,kavukcuoglu2026gemini4argon} have demonstrated the ability to perform robotic manipulation directly by generating actions from visual observations and adjusting them based on execution feedback. To support this direct use of general-purpose models, embodied harnesses provide interaction interfaces and structured context that help models apply their existing capabilities to specific robotic tasks. Show-Harness~\cite{chen2026show} provides a fine-grained action interface with clear semantics. RoboDawn~\cite{guo2026robodawn}, RoboICL~\cite{liu2026roboicl} and ReTrace~\cite{yu2026retrace} leverage demonstrations and experience gathered during interaction to help models adapt to tasks. Harness VLA~\cite{zhang2026harnessvla} adopts a memory-guided agentic framework that coordinates frozen VLA and analytic control primitives for robotic manipulation. These methods structure action interfaces and interaction context to support robotic manipulation.

\subsection{Real-to-Sim and Digital Twins}
Real-to-sim research constructs simulation environments from real-world observations for robot learning and interaction. NeRF~\cite{mildenhall2020nerf} and 3D Gaussian Splatting~\cite{kerbl20233dgaussiansplattingrealtime} provide the basis for reconstructing renderable scene representations and synthesizing novel views, while Re³Sim~\cite{han2026re3simgeneratinghighfidelitysimulation} combines mesh reconstruction with Gaussian-based rendering to construct high-fidelity simulation environments. More recently, Agentic Real2Sim~\cite{chen2026agentic} leverages vision-language agents to reconstruct simulation scenes and replay recorded robot interactions, while ARSTAG~\cite{li2026arstagagenticreal2sim2realtaskspecific}  employs language agents to construct task-specific simulation scenes and generate demonstrations at scale for policy learning. Beyond offline learning, reconstructed scenes are also used during deployment for planning, state tracking, and policy observations. Simify~\cite{kapelyukh2026test} reconstructs simulation-ready assets from a single RGB-D observation and uses evolutionary search with parallel physics rollouts at test time to jointly optimize object placement order and goal poses. GaussTwin~\cite{cai2026gausstwin} aims to unify simulation and online tracking of rigid bodies and deformable linear objects such as ropes, combining physics-based prediction with visual correction to maintain the interaction state; SyncTwin~\cite{huang2026synctwinfastdigitaltwin} targets safe manipulation in dynamic and partially occluded scenes, continuously updating object poses to provide planners with current scene geometry for collision avoidance and grasping. Real-is-Sim~\cite{abouchakra2025realissimbridgingsimtorealgap} uses a synchronized twin to maintain a consistent policy observation interface across training, simulation-based evaluation, and real-world deployment of learned policies. In SpatialHarness, the synchronized scene augments a frozen multimodal policy's observations with additional virtual views and structured spatial information during execution. \Cref{app:prior_sim} summarizes the capabilities of representative real-to-sim methods and presents a qualitative comparison with GaussTwin.

\begin{figure*}[t]
  \centering
  \makebox[\linewidth][c]{%
    \includegraphics[width=1.0\linewidth]{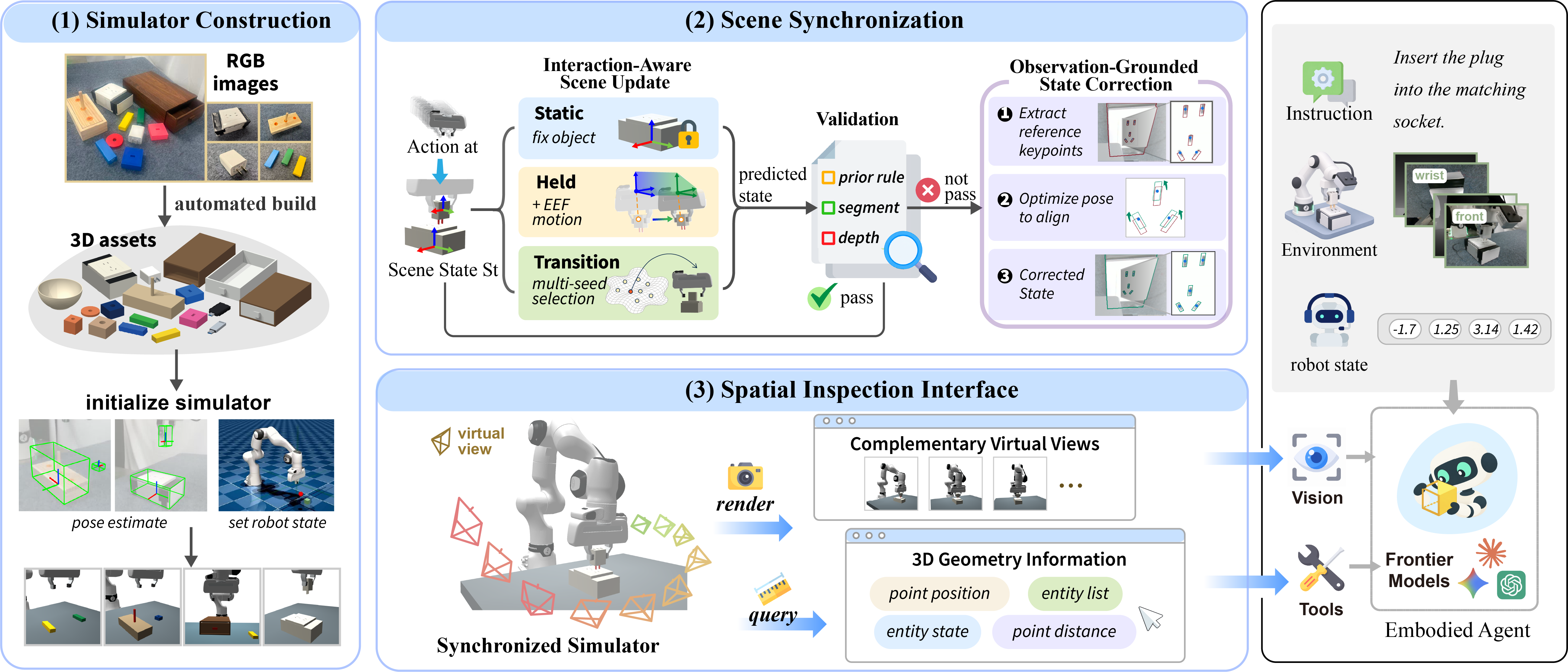}
  }
  \caption{\textbf{Simulator construction, scene synchronization, and spatial inspection.}
  (1) Reference images and object dimensions define the 3D models. Estimated object poses and the measured robot state initialize the simulator.
  (2) Static objects stay fixed, held objects follow the gripper, and the simulator replays grasps and releases from candidate object poses. It checks predicted poses against real RGB-D observations and corrects discrepancies.
  (3) The policy uses virtual views and geometric measurements, including keypoint positions and distances, alongside real observations to select its next action.}
  \label{fig:framework}
  \vspace{-15pt}
\end{figure*}

\section{Method}

We introduce SpatialHarness, a test-time spatial scaffolding framework that enhances the spatial observability of frozen policies for precise robotic manipulation. The key motivation is that successful manipulation often depends on precise spatial relationships that are difficult to infer from fixed physical camera views, especially when relevant geometry is occluded or poorly observed. SpatialHarness addresses this limitation by maintaining a synchronized simulated representation of the physical environment and exposing task-relevant spatial information through complementary virtual views and geometric queries. As illustrated in Fig.~\ref{fig:framework}, SpatialHarness comprises two components: (1) a Synchronized Simulator, which maintains an up-to-date geometric representation of the physical scene through interaction-aware state propagation and observation-grounded correction; and (2) Spatial Inspection Interface, which provide flexible virtual viewpoints and quantitative geometric measurements for spatial reasoning. These components are integrated into the existing robotic control loop to enhance the information available to the policy at inference time, without requiring policy fine-tuning or additional physical sensors.

\subsection{Problem Definition}
We consider VLM-based robotic manipulation, where a general-purpose VLM serves as a policy model or an action planner, guiding a robot $\mathcal{R}$ to execute an instruction $l$ in a real-world environment $\mathcal{E}$. Its execution process can be formulated as an observation–decision–execution loop. We index each iteration of the loop by timestep $t$, where $o_t$ denotes the real-world observations and $a_t$ denotes the action sequence produced by the policy. After $a_t$ is executed, the real environment transitions from $\mathcal{E}_{t}$ to $\mathcal{E}_{t+1}$. We denote the Synchronized Simulator by $\mathcal{S}$, with $s_t$ representing its simulated scene before execution of $a_t$. After $a_t$ is executed, the simulated scene is updated from $s_t$ to $s_{t+1}$ to reflect changes in the real environment. At timestep $t$, the policy can leverage existing observations $o_{1:t}$, combined with virtual views $v_t$ and spatial information $g_t$ obtained from simulated scene $s_t$ to make action decision:
\begin{equation}
a_t = \pi\left(l,o_{1:t},v_t,g_t\right),
\end{equation}
where $\pi$ denotes the policy model implemented by the VLM-based control framework.

\subsection{Synchronized Simulator}
The Synchronized Simulator maintains an explicit representation of the workspace, including object geometry, robot configuration, and spatial relationships. Unlike static scene reconstruction, the simulator continuously updates its state during physical interaction, enabling SpatialHarness to provide spatial context corresponding to the current execution state.

\subsubsection{Simulator Construction}

We first construct a simulation scene corresponding to the physical workspace. For each object, reference images and geometric dimensions are collected and provided to a VLM-based agent for CAD asset construction, as detailed in \Cref{app:asset_construction}. The agent also identifies task-relevant manipulation keypoints, such as the center of a drawer handle or socket opening, based on the task instruction and object affordances. These keypoints are defined in the local coordinate frame of each object, allowing their positions to be updated consistently with object motion. We then estimate the initial object poses from real-world observations and transform them into the robot base coordinate frame using calibrated camera extrinsics. The robot model and reconstructed objects are initialized in the simulator according to their robot state and estimated configurations, respectively. This establishes a shared geometric reference between the physical environment and its simulated counterpart, supporting subsequent scene synchronization and spatial inspection.

\subsubsection{Interaction-Aware Scene Synchronization}

Maintaining alignment between the simulated and physical environments is challenging during manipulation, as errors in object poses and contact modeling can cause simulated states to diverge from real execution. To address this, SpatialHarness combines simulation-based state propagation with observation-grounded correction. Given the current simulated scene $s_t$ and executed robot action $a_t$, the simulator predicts the next scene state, which is subsequently validated against real observations $o_{t+1}$ and corrected when necessary. To account for different interaction dynamics, we distinguish three object–gripper interaction modes: (1) \textit{Static mode}. For objects not involved in physical interaction, we preserve their previous poses during forward propagation and update them only when real observations indicate displacement. This avoids unnecessary simulation-induced errors. (2) \textit{Held mode}. For securely grasped objects, we propagate their poses according to the measured end-effector motion while maintaining the estimated object–gripper relationship. The resulting states are further checked against physical observations and gripper geometry to detect potential inconsistencies. (3) \textit{Transition mode}. Grasping and releasing involve changes in object–gripper contact relationships, making state propagation particularly sensitive to modeling errors. We therefore perform hypothesis-based interaction replay from a checkpoint before interaction. Specifically, we generate candidate states by perturbing object poses within geometrically constrained ranges and replay the executed robot motion from each candidate. The resulting states are evaluated against real observations using interaction-consistency checks and contour alignment, and the best-matching candidate is selected for subsequent refinement.

\subsubsection{Observation-Grounded State Correction}
Although interaction-aware propagation provides an estimate of the evolving scene, discrepancies may still arise from imperfect geometry, pose estimation, and physical modeling. SpatialHarness therefore validates the simulated state against real RGB-D observations and performs geometric correction when necessary. For each object, we first compare the predicted state with the segmentation masks and depth observations obtained from the external and wrist-mounted cameras. Predictions satisfying the observation-reliability criterion are accepted directly, avoiding unnecessary refinement. The reliability criterion requires that consistency with the observed segmentation masks and depth maps exceeds predefined thresholds and that geometric prior constraints are satisfied (e.g., the peg in Hanoi Tower should remain upright). Otherwise, the simulator triggers a correction procedure.
The correction begins with pose initialization from observed segmentation and depth information. We then refine the object's position and orientation by aligning its projected geometry with features extracted from the real observations. Specifically, object contours or task-relevant keypoints are extracted from the available camera views and used as references for geometric alignment target. The corresponding structures from the simulator are projected into the camera image plane, and the object pose is optimized to minimize the least-squares discrepancy between the projected and observed features. Geometric priors constrain the candidate poses and their adjustment ranges. For observation-based refinement, the external and wrist cameras provide complementary views, with the primary view selected according to observation quality. Pose refinement then uses geometry-specific features for alignment, such as the socket's hole pattern for insertion tasks.

\subsection{Spatial Inspection Interface}
Once the simulated scene is synchronized with the physical environment, SpatialHarness provides geometric representation through two complementary interfaces: virtual view rendering and geometric queries. These interfaces address different aspects of spatial reasoning. Virtual views make geometric relationships easier to perceive, while geometric queries provide explicit measurements for precise manipulation.

\noindent \textbf{Virtual view rendering.} Fixed physical cameras may not provide favorable perspectives for every manipulation task. For instance, the Hanoi Tower task requires precise alignment between the hole of the manipulated object and the target peg. Such relationships may be difficult to distinguish from the available camera views. SpatialHarness provides a \texttt{view\_scene} interface that renders the current synchronized scene from a specified virtual camera pose. By allowing the policy to inspect the scene from complementary viewpoints, this interface exposes spatial relationships that may be occluded or ambiguous in physical observations. 

\noindent \textbf{Geometric queries.} Although additional viewpoints improve visual access to the scene, precise manipulation may still require quantitative geometric information. SpatialHarness therefore provides structured interfaces for retrieving object geometry and spatial relationships directly from the simulator. Specifically, \texttt{list\_entities} enumerates scene objects and their associated manipulation keypoints. \texttt{get\_object\_size} returns object bounding-box information, while \texttt{get\_object\_pose} and \texttt{get\_point\_position} retrieve object poses and keypoint coordinates. The \texttt{get\_point\_distance} and \texttt{get\_angle} tools provide measurements of point separation and directional alignment, respectively. These tools enable the policy to reason explicitly about task-relevant geometric constraints, such as the distance between a plug and socket and the relative alignment of a hole and peg.

\subsection{Inference with SpatialHarness}

SpatialHarness integrates into existing VLM-based robotic control frameworks by augmenting the perception context available during action generation. The policy continues to receive real-world observations, task instructions, robot states, and execution feedback, while SpatialHarness provides additional spatial information from the synchronized simulator. We consider two integration paradigms: \textit{direct control}~\cite{gpt6astra, kavukcuoglu2026gemini4argon}, where the VLM generates robot actions directly, and \textit{agentic control}~\cite{zhang2026harnessvla}, where the VLM can invoke perception tools during planning. 

For direct control, SpatialHarness provides task-conditioned spatial information through virtual viewpoints. Before execution, a VLM analyzes the task instruction and initialized scene to identify task-critical spatial relationships and selects two complementary virtual viewpoints. At each decision step, the synchronized scene is rendered from these viewpoints, and object poses are retrieved from the simulator. These spatial observations are provided alongside real-world observations to support action generation. For agentic control, SpatialHarness exposes its spatial inspection interfaces as callable tools. The agent can dynamically request virtual views or geometric measurements based on its current information needs. For instance, it may inspect a drawer from a side view and query its opening distance to determine a collision-free placement action. The retrieved information is incorporated into the agent's perception context to guide subsequent decisions.

\section{Experiments}

\label{sec:th-experiments}

\begin{figure}[t]
  \centering
  \includegraphics[width=0.98\linewidth]{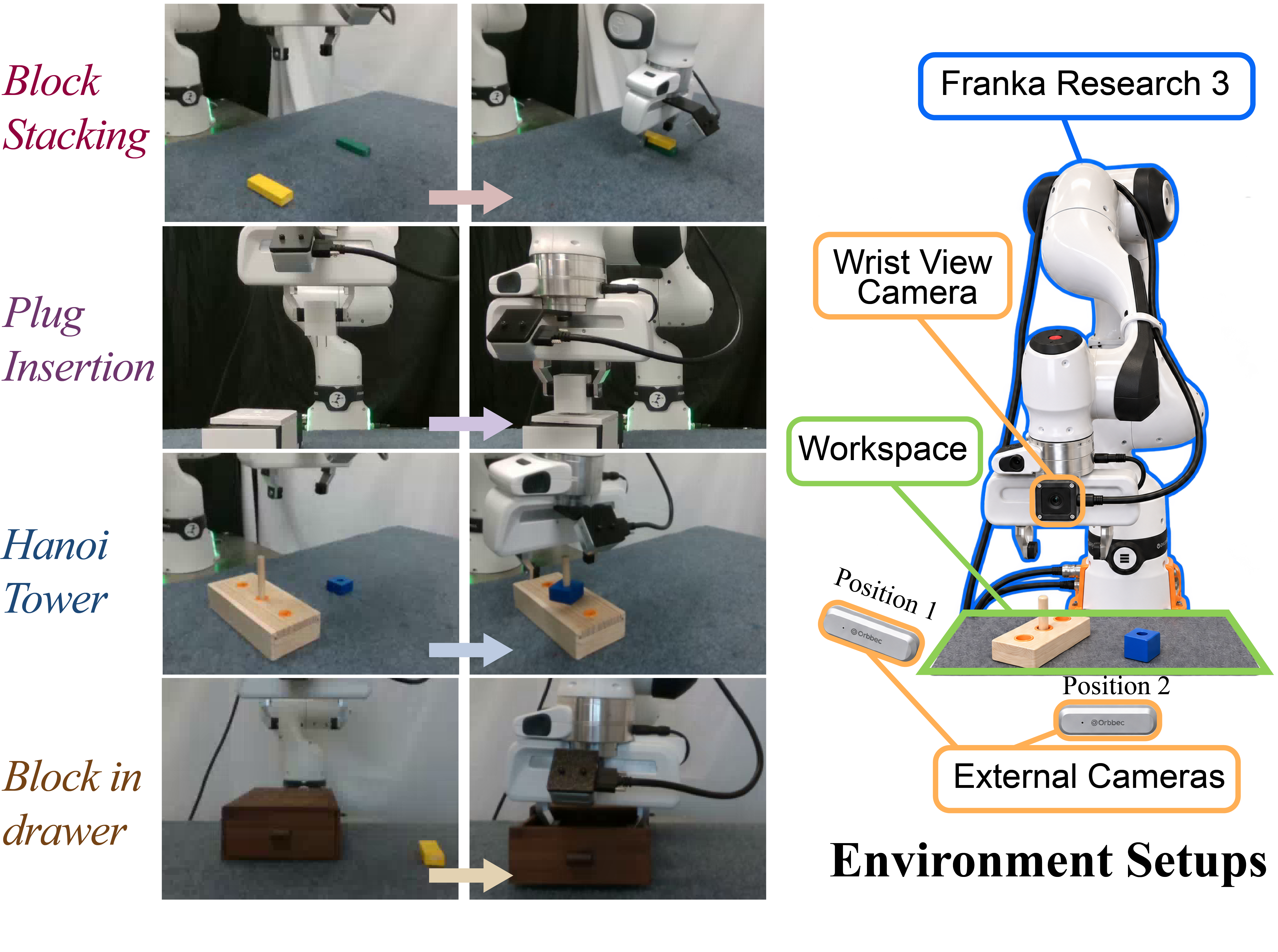}
  \caption{Real robot tasks and experimental platform. }
  \label{fig:demo}
  \vspace{-15pt}
\end{figure}

\subsection{Experimental Setup}
\noindent \textbf{Implementation Details.}
We use a Franka Research 3 arm and two Intel RealSense RGB-D cameras. One is mounted on the wrist, and the other provides a third-person view from one of two fixed external positions. MuJoCo~\cite{todorov2012mujoco} is used for physics simulation and image rendering. 

For direct control, we evaluate Robocurve's Inspect Robots framework~\cite{inspect-robots} with and without SpatialHarness using GPT-6.1 Sol~\cite{openai2026gpt61sol} with high reasoning effort, GPT-6 Astra~\cite{gpt6astra} with low reasoning effort, and Gemini-4 Argon with medium reasoning effort~\cite{kavukcuoglu2026gemini4argon}. For agentic control, our baseline is Harness VLA~\cite{zhang2026harnessvla}, an agentic embodied harness with segmentation- and depth-based perception tools. We augment it with our spatial inspection interfaces. Both variants use GPT-6.1 Sol (high reasoning effort) as the agentic planner and are evaluated without carrying over experience between trials. All policies receive the task instruction, RGB images from both real cameras, measured robot states, and execution feedback. SpatialHarness additionally provides complementary virtual views and spatial information from the Synchronized Simulator.

\noindent \textbf{Tasks.} \Cref{fig:demo} presents the four tasks:
\begin{itemize}
    \item \textbf{\textit{Block stacking}} requires grasping a block, aligning it above another
    block, and releasing it into a stable stack.
    \item \textbf{\textit{Plug insertion}} requires matching the position and orientation of
    a plug to its socket before insertion.
    \item \textbf{\textit{Hanoi Tower}} evaluates placement of a holed piece over a target
    peg, where the policy must assess the relative alignment of the hole and peg.
    \item \textbf{\textit{Block in Drawer}} requires opening an articulated drawer
    and then placing a block inside.
\end{itemize} 

\begin{table*}[t]
\centering
\definecolor{SHMainRowGray}{HTML}{EFEFEF}
\caption{Task success rates (\%) under direct and agentic control.
Average denotes the mean success rate across the four tasks.
Green arrows indicate absolute improvements in percentage points
over the corresponding baselines.}
\label{tab:overall_success}

\fontsize{8}{9.5}\selectfont
\renewcommand{\arraystretch}{1.0}
\setlength{\extrarowheight}{0pt}
\setlength{\tabcolsep}{3.5pt}

\newcommand{\resultrowstrut}{\rule[-1.25ex]{0pt}{3.7ex}}

\newcommand{\resultgain}[2]{%
    \mbox{%
        \textbf{#1\%}\hspace{0.08em}%
        {\fontsize{6.5}{7.8}\selectfont
        \textcolor{green!50!black}{(\ensuremath{\uparrow}#2)}}%
    }%
}

\newcommand{\resultlogo}[1]{%
    \raisebox{-0.25ex}{%
        \includegraphics[height=1.75ex]{#1}%
    }%
    \hspace{0.4em}%
}

\newcommand{\resultbaseline}[2]{%
    \resultrowstrut
    \resultlogo{#1}%
    #2%
}

\newcommand{\resultours}{%
    \resultrowstrut
    \hspace*{2em}+ \textbf{SpatialHarness (Ours)}%
}

\begin{tabularx}{\textwidth}{
    l
    *{5}{>{\centering\arraybackslash}X}
}
\specialrule{\heavyrulewidth}{0pt}{0pt}

\rule[-1.225ex]{0pt}{3.65ex}\textbf{Method}
& \mbox{\textbf{Block Stacking}}
& \mbox{\textbf{Plug Insertion}}
& \mbox{\textbf{Hanoi Tower}}
& \mbox{\textbf{Block in Drawer}}
& \textbf{Average} \\

\specialrule{\lightrulewidth}{0pt}{0pt}

\rowcolor{WAMGreen}
\multicolumn{6}{c}{%
    \rule[-1.125ex]{0pt}{3.45ex}\textbf{\textit{Direct Control}}} \\

\specialrule{\lightrulewidth}{0pt}{0pt}

\resultbaseline{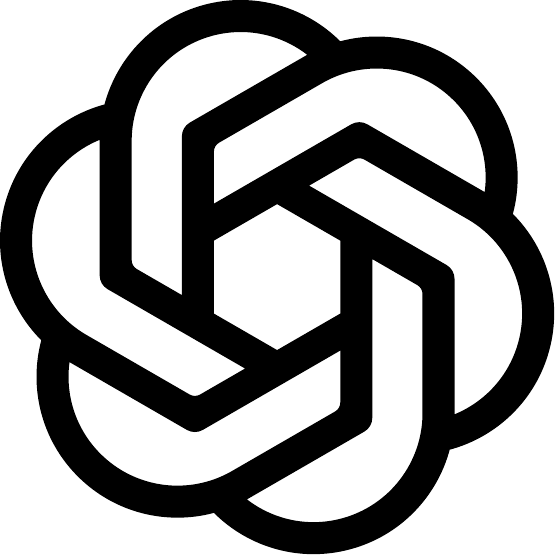}{GPT-6.1 Sol high}
& 40.0\%
& 13.3\%
& 0.0\%
& 0.0\%
& 13.3\% \\

\rowcolor{SHMainRowGray}
\resultours
& \resultgain{100.0}{60.0}
& \resultgain{80.0}{66.7}
& \resultgain{6.7}{6.7}
& \resultgain{60.0}{60.0}
& \resultgain{61.7}{48.3} \\

\specialrule{\lightrulewidth}{0pt}{0pt}

\resultbaseline{fig/gpt_logo.pdf}{GPT-6 Astra low}
& 46.7\%
& 26.7\%
& 0.0\%
& 0.0\%
& 18.3\% \\

\rowcolor{SHMainRowGray}
\resultours
& \resultgain{100.0}{53.3}
& \resultgain{66.7}{40.0}
& \resultgain{100.0}{100.0}
& \resultgain{66.7}{66.7}
& \resultgain{83.3}{65.0} \\

\specialrule{\lightrulewidth}{0pt}{0pt}

\resultbaseline{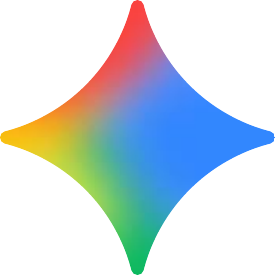}{Gemini-4 Argon medium}
& 80.0\%
& 33.3\%
& 0.0\%
& 0.0\%
& 28.3\% \\

\rowcolor{SHMainRowGray}
\resultours
& \resultgain{93.3}{13.3}
& \resultgain{66.7}{33.3}
& \resultgain{73.3}{73.3}
& \resultgain{73.3}{73.3}
& \resultgain{76.7}{48.3} \\

\specialrule{\lightrulewidth}{0pt}{0pt}

\rowcolor{VLABlue}
\multicolumn{6}{c}{%
    \rule[-1.125ex]{0pt}{3.45ex}\textbf{\textit{Agentic Control}}} \\

\specialrule{\lightrulewidth}{0pt}{0pt}

\resultbaseline{fig/gpt_logo.pdf}{Harness VLA (GPT-6.1 Sol high)}
& 100.0\%
& 86.7\%
& 60.0\%
& 46.7\%
& 73.3\% \\

\rowcolor{SHMainRowGray}
\resultours
& \textbf{100.0\%}
& \resultgain{100.0}{13.3}
& \resultgain{100.0}{40.0}
& \resultgain{86.7}{40.0}
& \resultgain{96.7}{23.3} \\

\specialrule{\heavyrulewidth}{0pt}{0pt}
\end{tabularx}
\end{table*}

\noindent \textbf{Metrics.} We report task success rates over 15 trials for Real-world experiments and both the Spatial Inspection Interface and Scene Synchronization ablations, and over 10 trials for the thinking effort ablation. Each trial is limited to 20 dialogue turns under direct control and to 20 action steps and 50 dialogue turns under agentic control. Commands of different types issued in the same decision, such as translation and rotation, count as one action step. Each trial ends upon task completion or when a budget limit is reached. \Cref{tab:success_criteria} in Appendix lists the success criteria for all four tasks.

\begin{figure*}[t]
  \centering
  \makebox[\linewidth][c]{%
    \includegraphics[width=0.80\linewidth]{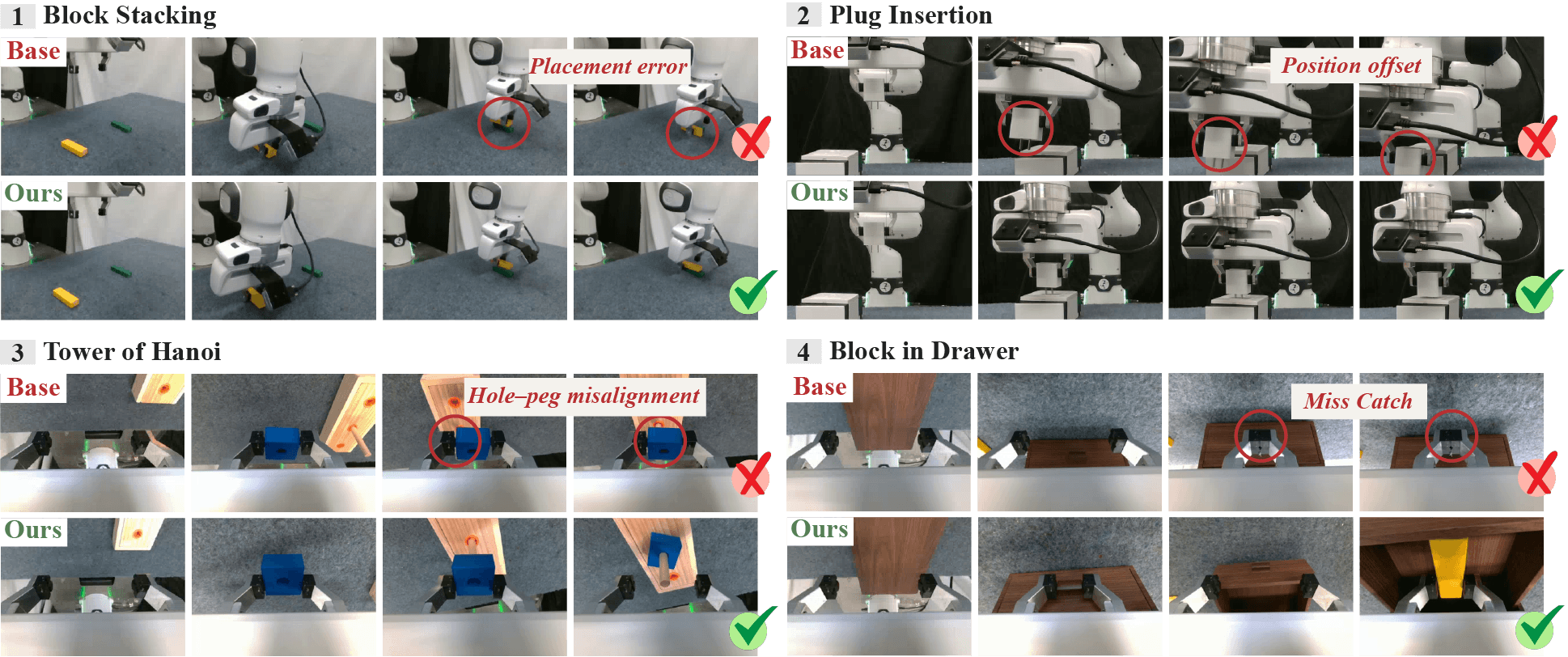}
  }
  \caption{Representative direct-control trials with GPT-6 Astra. In each task panel, frames run from left to right, with the baseline above and the same policy with SpatialHarness below. Red crosses and green checkmarks indicate failure and success in the trials shown.}
  \label{fig:visualization}
  \vspace{-15pt}
\end{figure*}

\begin{figure}[t]
  \centering
  \makebox[\linewidth][c]{%
    \includegraphics[width=1.0\linewidth]{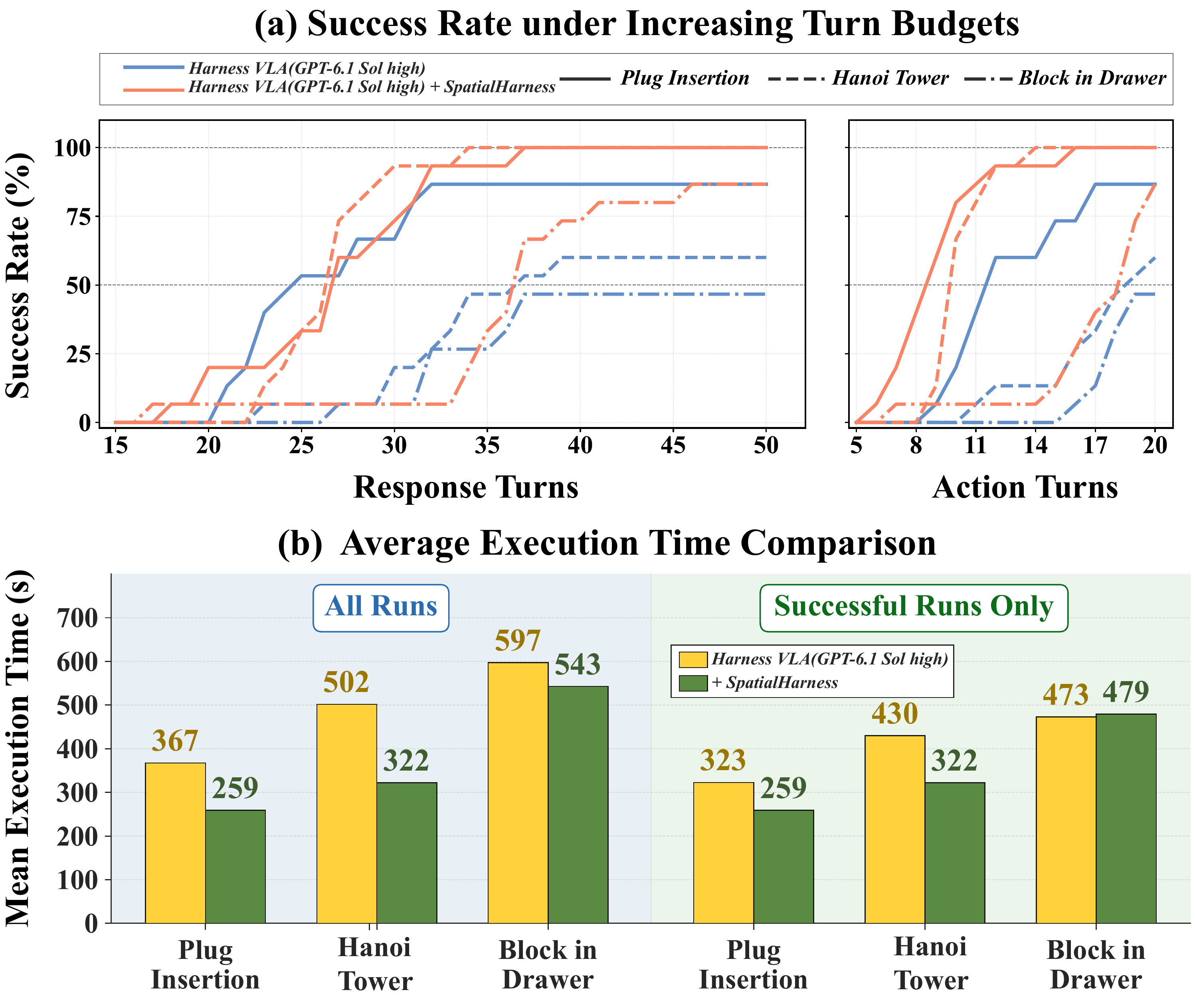}
  }
  \caption{Efficiency comparison on three tasks. (a) Task success rates under increasing execution budgets. The left plot tracks model responses received by the planner, while the right tracks robot action commands. Success rate is the percentage of all trials succeeding within budget. (b) Average execution time for all trials and successful trials only. Timing starts when the planner first calls the model and ends when the task finishes or the execution budget is exhausted.}
  \label{fig:efficiency}
  \vspace{-18pt}
\end{figure}

\subsection{Real-world Experiments}
To evaluate SpatialHarness across different models and control frameworks, we compare each configuration with and without SpatialHarness, with the latter serving as the baseline. As shown in \Cref{tab:overall_success}, we conduct experiments on four real-robot manipulation tasks using three closed-source VLMs for direct control and Harness VLA with GPT-6.1 Sol high as its planner for agentic control.

\noindent \textbf{Under direct control, SpatialHarness substantially improves task success rates across all evaluated models.} Compared with their baselines under direct control, SpatialHarness improves the average success rates of GPT-6.1 Sol high, GPT-6 Astra low, and Gemini-4 Argon medium by 48.3, 65.0, and 48.3 percentage points, respectively. Notably, SpatialHarness enables both GPT-6 Astra low and Gemini-4 Argon medium to complete Hanoi Tower and Block in Drawer in the majority of trials, although neither model succeeds on either task without it. As shown in \Cref{fig:visualization}, GPT-6 Astra low fails to align the hole with the peg due to gripper occlusion and misses the drawer handle in the respective baseline trials. With complementary virtual views and explicit object poses from the synchronized simulator, the same policy overcomes both difficulties.

\noindent \textbf{More accurate spatial information does not necessarily translate into successful manipulation.} 
With SpatialHarness, GPT-6.1 Sol high slightly outperforms GPT-6 Astra low on Plug Insertion, but achieves only 6.7\% on Hanoi Tower, compared to GPT-6 Astra low's 100.0\%. 
Execution traces reveal different manipulation strategies that may explain this contrast. 
GPT-6.1 Sol high tends to attempt insertion before sufficient alignment. The socket's sloped edges can guide the plug into place, but the Hanoi peg offers no such guidance, making premature insertion more likely to fail.
In contrast, GPT-6 Astra low iteratively refines manipulation poses, enabling precise alignment in Hanoi Tower but sometimes exhausting interaction budgets on Plug Insertion.

\noindent \textbf{SpatialHarness provides additional spatial information beyond existing perception tools for agentic control framework.} 
Although Harness VLA is already equipped with segmentation-based and depth-based perception tools, integrating SpatialHarness further increases its average success rate from 73.3\% to 96.7\%, with notable gains from 60.0\% to 100.0\% on Hanoi Tower and from 46.7\% to 86.7\% on Block in Drawer. Through virtual views and explicit geometric queries, SpatialHarness offers access to spatial relationships that are difficult to infer from segmentation and depth observations alone. These results demonstrate the value of simulation-based spatial information beyond conventional perception tools.

\noindent \textbf{Efficiency.}
We compare the efficiency of Harness VLA and our method on Plug Insertion, Hanoi Tower, and Block in Drawer. As shown in panel (a) of \Cref{fig:efficiency}, our method reaches the baseline’s final success rate with 6 and 10 fewer action turns on Plug Insertion and Hanoi Tower, respectively. Under the same action budgets, it also achieves higher success rates on all three tasks. For the model response budget, SpatialHarness achieves higher final success rates than the baseline on all three tasks, with a clear advantage on Hanoi Tower even at lower budgets. As shown in panel (b) of \Cref{fig:efficiency}, our method reduces the average execution time across all trials by 29.4\%, 35.9\%, and 9.0\% on the three tasks, respectively. For successful trials, the average completion times for Plug Insertion and Hanoi Tower still decrease by 19.8\% and 25.1\%. This indicates that the time advantage also holds for successful runs. For Block in Drawer, the average completion time for successful trials changes only slightly from 473 s to 479 s, while the success rate increases from 46.7\% to 86.7\%, indicating that our method achieves a higher success rate with a comparable average completion time. These results show that our method improves both task success rates and execution efficiency.

\begin{table*}[!t]
\providecolor{WAMGreen}{RGB}{241,248,243}
\providecolor{VLABlue}{RGB}{241,246,250}

\newcommand{\SHMethodCell}[2]{%
  \parbox[c][#1][c]{\linewidth}{\raggedright #2\par}}
\newcommand{\SHScoreCell}[2]{%
  \parbox[c][#1][c]{\linewidth}{\centering #2\par}}
\newcommand{\SHGain}[2]{%
  \mbox{#1\hspace{0.08em}{\fontsize{6.5}{7.8}\selectfont
    \textcolor{green!50!black}{(\ensuremath{\uparrow}#2)}}}}
\newcommand{\SHLogo}{%
  \raisebox{-0.25ex}{\includegraphics[height=1.75ex]{fig/gpt_logo.pdf}}%
  \hspace{0.4em}}
\newcommand{\SHRule}{\specialrule{\lightrulewidth}{0pt}{0pt}}
\newcommand{\SHHeavyRule}{\specialrule{\heavyrulewidth}{0pt}{0pt}}
\newcommand{\SHControl}[2]{%
  \rowcolor{#1}\multicolumn{4}{c}{\rule[-1.05ex]{0pt}{3.3ex}%
    \textbf{\textit{#2}}}\\\SHRule}
\newcommand{\SHScores}[4]{%
  \SHScoreCell{#1}{#2} & \SHScoreCell{#1}{#3} & \SHScoreCell{#1}{#4}}
\newcommand{\SHComponentIndent}{\hspace*{0.6em}}
\newcommand{\SHFullMethod}[2]{%
  \setlength{\baselineskip}{8.7pt}%
  \SHComponentIndent+ #1\par
  \SHComponentIndent\mbox{+ #2~\textbf{(full)}}}

\newcommand{\SHAblationContents}[6]{%
  \centering
  \caption{#6}
  \label{#1}
  \fontsize{7.5}{9}\selectfont
  \renewcommand{\arraystretch}{1.0}
  \setlength{\extrarowheight}{0pt}
  \setlength{\tabcolsep}{0.6pt}
  \newcommand{\SHSingleHeight}{3.5ex}
  \newcommand{\SHAgentHeight}{6.2ex}
  \newcommand{\SHFullHeight}{6.2ex}
  \begin{tabularx}{\columnwidth}{@{}>{\raggedright\arraybackslash}X
    *{3}{>{\centering\arraybackslash}p{0.21\columnwidth}}@{}}
  \SHHeavyRule
  \SHMethodCell{3.6ex}{\strut\textbf{Method}} &
  \SHScores{3.6ex}{\strut\textbf{Plug Insertion}}
    {\strut\textbf{Hanoi Tower}}{\strut\textbf{Average}}\\
  \SHRule
  \SHControl{WAMGreen}{Direct Control}
  \SHMethodCell{\SHSingleHeight}{\SHLogo GPT-6.1 Sol high} &
  \SHScores{\SHSingleHeight}{13.3\%}{0.0\%}{6.7\%}\\
  \SHRule
  \SHMethodCell{\SHSingleHeight}{\SHComponentIndent+ #2} & #4\\
  \SHRule
  \rowcolor{gray!8}
  \SHMethodCell{\SHFullHeight}{\SHFullMethod{#2}{#3}} &
  \SHScores{\SHFullHeight}{\SHGain{\textbf{80.0\%}}{66.7}}
    {\SHGain{\textbf{6.7\%}}{6.7}}{\SHGain{\textbf{43.3\%}}{36.7}}\\
  \SHRule
  \SHControl{VLABlue}{Agentic Control}
  \SHMethodCell{\SHAgentHeight}{\SHLogo Harness VLA\par
    \hspace*{0.75em}(GPT-6.1 Sol high)} &
  \SHScores{\SHAgentHeight}{86.7\%}{60.0\%}{73.3\%}\\
  \SHRule
  \SHMethodCell{\SHSingleHeight}{\SHComponentIndent+ #2} & #5\\
  \SHRule
  \rowcolor{gray!8}
  \SHMethodCell{\SHFullHeight}{\SHFullMethod{#2}{#3}} &
  \SHScores{\SHFullHeight}{\SHGain{\textbf{100.0\%}}{13.3}}
    {\SHGain{\textbf{100.0\%}}{40.0}}{\SHGain{\textbf{100.0\%}}{26.7}}\\
  \SHHeavyRule
  \end{tabularx}}

\newcommand{\SHSpatialInterfaceContents}{%
  \SHAblationContents{tab:ablation}{Geometry Queries}{Virtual Rendering}
    {\SHScores{\SHSingleHeight}{\SHGain{40.0\%}{26.7}}{0.0\%}
      {\SHGain{20.0\%}{13.3}}}
    {\SHScores{\SHSingleHeight}{\SHGain{100.0\%}{13.3}}
      {\SHGain{66.7\%}{6.7}}{\SHGain{83.3\%}{10.0}}}
    {Ablation of the Spatial Inspection Interface.}}

\newcommand{\SHSceneSyncContents}{%
  \SHAblationContents{tab:scene_sync_ablation}{State Propagation}{State Correction}
    {\SHScores{\SHSingleHeight}{\SHGain{66.7\%}{53.3}}{0.0\%}
      {\SHGain{33.3\%}{26.7}}}
    {\SHScores{\SHSingleHeight}{\SHGain{100.0\%}{13.3}}
      {\SHGain{86.7\%}{26.7}}{\SHGain{93.3\%}{20.0}}}
    {Ablation of Scene Synchronization.}}

  \centering
  \begin{minipage}[t]{\dimexpr(\textwidth-\columnsep)/2\relax}
    \vspace{0pt}\SHSpatialInterfaceContents
  \end{minipage}\hfill
  \begin{minipage}[t]{\dimexpr(\textwidth-\columnsep)/2\relax}
    \vspace{0pt}\SHSceneSyncContents
  \end{minipage}
\end{table*}

\subsection{Ablation Studies}

We conduct ablation studies on Plug Insertion and Hanoi Tower under direct and agentic control, using GPT-6.1 Sol and Harness VLA~\cite{zhang2026harnessvla} as the respective baselines. In these two ablations, GPT-6.1 Sol high serves as the policy model under direct control and as the planner in Harness VLA.

\noindent \textbf{Spatial Inspection Interface.}
\Cref{tab:ablation} compares the baselines, their variants with Geometry Query interfaces, and full SpatialHarness. In the Geometry Queries variant, the policy receives all object poses as input under direct control and accesses geometric information through a callable query interface under agentic control. Full SpatialHarness additionally enables virtual view rendering. Under direct control, the policy lacks an  an explicit tool-based spatial reasoning process and appears to benefit more from visual spatial information. On Plug Insertion, the Geometry Queries variant improves success by 26.7 percentage points to 40.0\%, whereas adding virtual view rendering further raises it to 80.0\%. However, direct control achieves only 6.7\% on Hanoi Tower even with full SpatialHarness, suggesting that the policy's reasoning and control capabilities remain a limiting factor. Under agentic control, the benefits of virtual views depend on the task's spatial requirements. On Plug Insertion, the Geometry Queries variant increases success from 86.7\% for the baseline to 100.0\%, with no further improvement from virtual view rendering. This suggests that the policy can use explicit geometric measurements to reason about the relative spatial relationship between the plug and socket. On Hanoi Tower, however, the critical peg--hole alignment is obscured in real observations. Although geometric queries return object poses, they do not capture the local boundaries of the peg and hole needed to assess their precise alignment. The Geometry Queries variant therefore increases success only from 60.0\% for the baseline to 66.7\%, whereas adding virtual view rendering allows the policy to visually inspect this alignment from an overhead view and further raises success to 100.0\%. These findings suggest that the effect of spatial information depends on both how it is presented and the control framework's ability to interpret it.

\noindent \textbf{Scene Synchronization.} \Cref{tab:scene_sync_ablation} compares the baselines, their State Propagation variants with spatial inspection, and full SpatialHarness. In the State Propagation variant, the simulated scene is aligned with real RGB-D observations at the start of each trial and updated only through state propagation during execution. This variant retains observation-guided candidate selection during grasp and release, but uses the propagated pose even when it fails the alignment check. Full SpatialHarness adds online State Correction to this variant, so the policy receives spatial information from a scene refined to align with real observations. As shown in \Cref{tab:scene_sync_ablation}, the State Propagation variant raises mean success across the two tasks from 6.7\% to 33.3\% under direct control and from 73.3\% to 93.3\% under agentic control. These gains show that the propagated scene from an aligned initial state can provide useful spatial information for control. Adding online State Correction further raises these averages to 43.3\% and 100.0\%, respectively. Online correction helps reduce the impact of residual pose errors on alignment decisions. On Hanoi Tower, it raises success under agentic control from 86.7\% to 100.0\%. Here, the hole and peg may appear aligned in simulation but remain offset in the real scene, misleading the policy. Correction updates the simulated poses, giving the policy spatial information consistent with reality. On Plug Insertion, the socket's sloped edges can guide a slightly misaligned plug into place, which may explain the high success rate achieved with State Propagation alone under agentic control.

\begin{figure}[t]
  \centering
  \makebox[\linewidth][c]{%
    \includegraphics[width=0.96\linewidth]{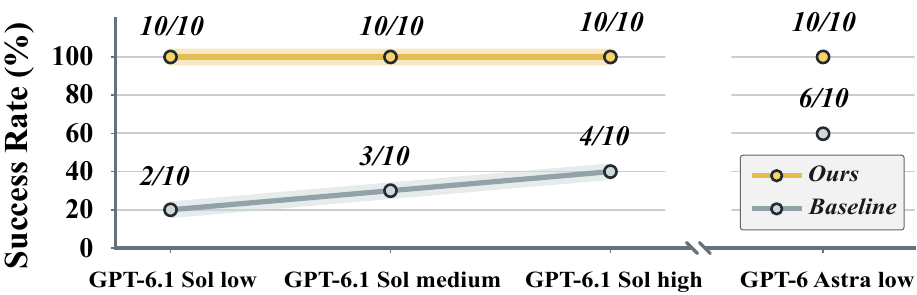}
  }
  \caption{Effect of model capability and reasoning effort.}
  \label{fig:thingking_effort}
  \vspace{-18pt}
\end{figure}

\noindent \textbf{Ablation of model thinking effort and capacity.}
We further examine the effectiveness of our method across models and reasoning effort levels. To reduce the influence of harness mechanisms, we evaluate both methods on the block stacking task under direct control, with 10 trials per configuration. For configurations also evaluated in the main experiments, we use the first 10 trials from their corresponding 15-trial evaluations. As shown in \Cref{fig:thingking_effort}, increasing GPT-6.1 Sol's reasoning effort from low to high raises the baseline success rate from 20\% to 40\%, while switching to GPT-6 Astra at low reasoning effort raises it to 60\%. With SpatialHarness, all four configurations succeed in 10 of 10 trials.

\section{Conclusion and Limitation}
\noindent \textbf{Conclusion.} We have presented SpatialHarness, a test-time embodied harness that improves spatial observability for fine robotic manipulation. Through interaction-aware scene synchronization and task-oriented view selection, it provides frozen multimodal policies with spatial context comprising complementary virtual views and structured spatial information. Real-robot experiments show substantially improved manipulation success without policy fine-tuning or changes to physical cameras, highlighting how test-time spatial scaffolding can help realize existing model capabilities.
\noindent \textbf{Limitation.} SpatialHarness currently leverages frontier multimodal foundation models, whose closed-source nature and inference latency may constrain deployment efficiency. However, continued advances in smaller open-source models offer promising opportunities for more accessible and efficient implementations. Additionally, SpatialHarness relies on a sufficiently accurate simulated environment, where imperfect geometry and pose estimation may affect spatial information quality. While our interaction-aware synchronization mitigates such discrepancies, more automated scene construction and efficient synchronization could further improve scalability to diverse and dynamic environments. Future work will explore these directions to enable broader and more efficient deployment of SpatialHarness.

{
    \small
    \bibliographystyle{ieeenat_fullname}
    \bibliography{main}
}

\clearpage
\appendix
\section{Study with an Open-Source Model}
To explore whether SpatialHarness can reduce reliance on proprietary models, we evaluate it with the open-source multimodal foundation model DeepSeek-V4.1-Flash-1017~\cite{deepseekai2026deepseekv41flash} at maximum reasoning effort. Initial trials with this model under direct control showed little progress toward task completion, so we conduct the evaluation under agentic control.
As in the main experiments, timing starts with the planner's first model call. For a successful trial, timing ends when the robot finishes the action that we judge to have met the physical success criterion in \Cref{tab:success_criteria}. For an unsuccessful trial, timing ends when the execution budget is exhausted.

As shown in \Cref{tab:deepseek}, SpatialHarness improves the success rate on the Plug Insertion task from 13.3\% to 73.3\%, an absolute gain of 60.0 percentage points. All 15 baseline trials used the full 50 response turns. In both successful baseline trials, the final robot action inserted the plug into the socket. SpatialHarness averaged 39.3 response turns. Average execution time also decreases from 651 s to 353 s, a reduction of 45.8\%. These results provide initial evidence that SpatialHarness can help an open-source model find successful task trajectories more efficiently under the same evaluation budget. They also demonstrate that SpatialHarness is applicable to an open-source model and can substantially improve its manipulation performance.

\section{3D Asset Construction}
\label{app:asset_construction}

\begin{figure}[t]
  \centering
  \makebox[\linewidth][c]{%
    \includegraphics[width=\linewidth]{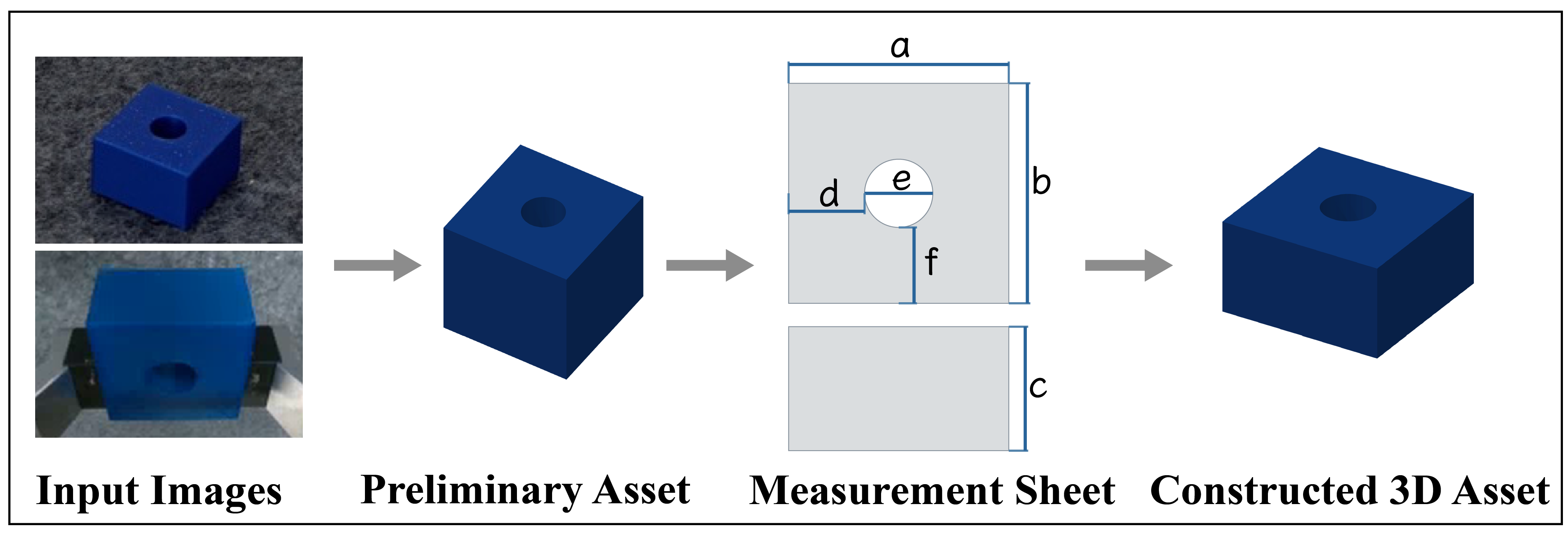}
  }
  \caption{Visualization of the CAD modeling process}
  \label{fig:cad_construction}
\end{figure}

We formalize the CAD modeling process as a skill consisting of four stages: preliminary asset construction, geometric dimension annotation, model refinement and alignment, and export and validation, as shown in \Cref{fig:cad_construction}. The goal is to produce a CAD model that accurately captures the physical object’s geometry and dimensions for robotic manipulation.

\noindent \textbf{1. Preliminary asset construction.} The CAD reconstruction process begins with 2–4 input images of the same object taken from different viewpoints. The images should show the overall shape, side surfaces, and important features such as holes and slots. Based on the images, a multimodal agent identifies the object’s geometric features and represents its structure using templates or basic shapes. For example, a block from the Hanoi Tower can be modeled as a rectangular prism with a cylindrical through-hole. Its length, width, thickness, hole diameter, and hole position are defined as adjustable parameters. This forms a preliminary asset that captures the object’s structure, while its dimensions remain provisional.

\noindent \textbf{2. Geometric dimension annotation.} The agent generates orthographic views of the preliminary asset and uses dimension lines and labels to indicate the measurements needed for reconstruction. These include the object’s overall dimensions, the sizes of local features, and their positions relative to the object’s edges. For example, the top view of a Tower of Hanoi block can show its length, width, hole diameter, and the distances from the hole to the edges, while a side view shows its thickness. The operator measures the physical object according to these annotations and returns the results to the agent in millimeters.

\noindent \textbf{3. Model refinement and alignment.} The agent updates the model’s geometric parameters using the measured values and checks their consistency with the model’s geometric constraints. For example, if the block’s through-hole is centered, its distances to the two opposite edges should be equal, and those distances plus the hole diameter should match the block’s overall width. Any discrepancies are reported for review and correction.

\noindent \textbf{4. Export and validation.} Once the structure and dimensions are confirmed, the agent generates and exports the final CAD model.

\begin{table}[t]
\centering
\caption{Performance of SpatialHarness with
DeepSeek-V4.1-Flash-1017 Max under agentic control.
Green arrows indicate absolute improvements in
percentage points over the baseline. Response turns and execution
times are averages over all 15 trials.}
\label{tab:deepseek}

\small
\renewcommand{\arraystretch}{1.0}
\setlength{\extrarowheight}{0pt}
\setlength{\tabcolsep}{2pt}

\begin{NoHyper}
\begin{tabularx}{\columnwidth}{@{}
    >{\raggedright\arraybackslash}X
    >{\centering\arraybackslash}p{0.21\columnwidth}
    >{\centering\arraybackslash}p{0.12\columnwidth}
    >{\centering\arraybackslash}p{0.14\columnwidth}
@{}}

\specialrule{\heavyrulewidth}{0pt}{0pt}

\multirow{2}{*}[-1.5ex]{\textbf{Method}}
& \multicolumn{3}{c}{\rule{0pt}{2.9ex}\textbf{Task -  Plug Insertion}} \\
\cmidrule(lr){2-4}
\rule[-2.7ex]{0pt}{0pt}
& \parbox[t]{\linewidth}{\centering\textbf{Succ.}\\\textbf{Rate}}
& \parbox[t]{\linewidth}{\centering\textbf{Resp.}\\\textbf{Turn}}
& \parbox[t]{\linewidth}{\centering\textbf{Exec.}\\\textbf{Time}} \\

\specialrule{\lightrulewidth}{1.2ex}{0pt}

\parbox[c]{\linewidth}{%
    \vspace{0.6ex}
    \raisebox{-0.25ex}{%
        \includegraphics[height=2ex]{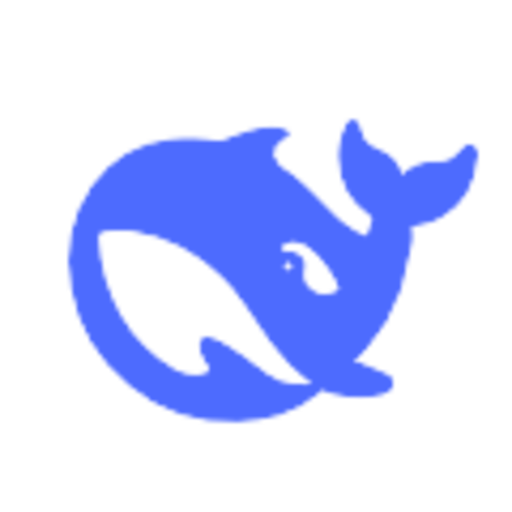}%
    }%
    \hspace{0.4em}Harness VLA~\cite{zhang2026harnessvla}\par
    \hspace*{0.5em}{\scriptsize (DeepSeek-V4.1-Flash-1017 Max~\cite{deepseekai2026deepseekv41flash})}%
    \vspace{0.6ex}
}
& 13.3\% & 50.0 & 651s \\

\specialrule{\lightrulewidth}{0pt}{0pt}

\rowcolor{gray!8}
\rule[-1.4ex]{0pt}{4.2ex}
\hspace*{0.4em}+ SpatialHarness (Ours)
&
\mbox{%
    \textbf{73.3\%}\hspace{0.2em}%
    {\fontsize{6.5}{7.5}\selectfont
    \textcolor{green!50!black}{%
        ($\uparrow$60.0)%
    }}%
} & 39.3 & 353s \\

\specialrule{\heavyrulewidth}{0pt}{0pt}

\end{tabularx}
\end{NoHyper}
\end{table}

\section{Comparison with Prior Simulation Methods}
\label{app:prior_sim}

\begin{figure*}[t]
  \centering
  \includegraphics[width=0.88\linewidth]{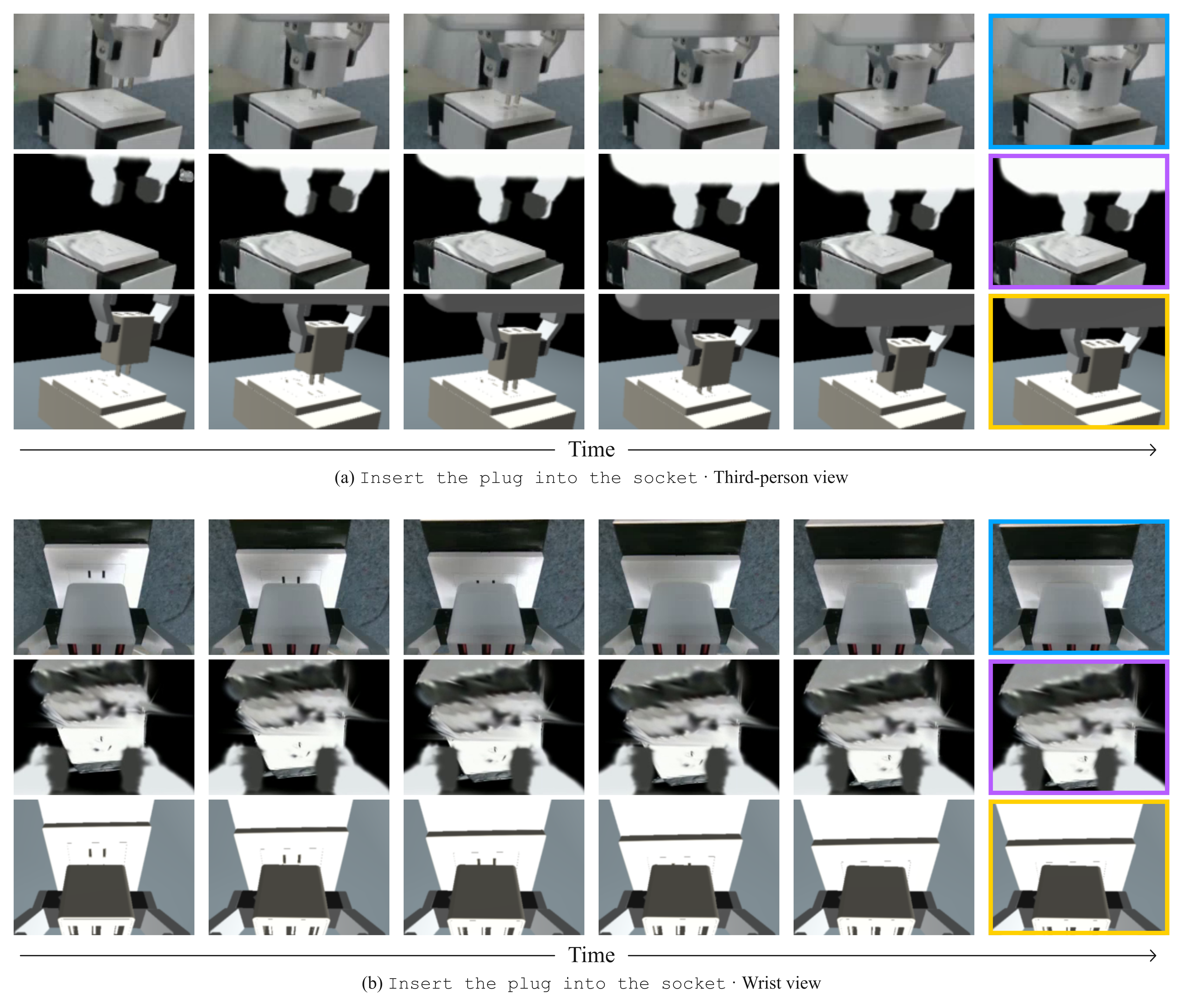}
  \caption{Plug insertion from (a) the external camera and (b) the wrist camera. Each panel shows six time points from left to right. In the final column, colored borders identify the \textcolor{ObservationBlue}{real observation (top)}, the \textcolor{GaussTwinPurple}{GaussTwin rendering (middle)}, and the \textcolor{SimulationYellow}{SpatialHarness rendering (bottom)}.}
  \label{fig:gausstwin_plug_insertion}
\end{figure*}

\begin{table*}[t]
\centering
\caption{Capabilities and public code availability of real-to-sim methods.}
\label{tab:prior_simulation_methods}

\fontsize{9}{9.5}\selectfont
\setlength{\tabcolsep}{2pt}
\renewcommand{\arraystretch}{1.1}

\begin{NoHyper}
\begin{tabular*}{0.95\linewidth}{@{\extracolsep{\fill}}l*{7}{c}@{\hspace{8pt}}}
\toprule
\textbf{Attribute}
& Re\textsuperscript{3}Sim
& Simify
& SyncTwin
& Real-is-Sim
& Agentic Real2Sim
& GaussTwin
& \textbf{Ours} \\
\midrule
Publication Venue
& ICRA 2026 & IROS 2026 & RA-L 2026 & ICRA 2026
& CoRL 2026 Workshop & ICRA 2026 & \\
\midrule
Training-Free Construction
& & \ding{51} & \ding{51} & & \ding{51} & & \ding{51} \\
Initial Scene Alignment
& & \ding{51} & \ding{51} & \ding{51} & \ding{51} & \ding{51} & \ding{51} \\
Online Synchronization
& & & \ding{51} & \ding{51} & & \ding{51} & \ding{51} \\
Additional Virtual Views
& \ding{51} & & & \ding{51} & \ding{51} & \ding{51} & \ding{51} \\
Geometry Access
& \ding{51} & \ding{51} & \ding{51} & \ding{51} & \ding{51} & \ding{51} & \ding{51} \\
\midrule
Code Release
& \ding{51} & & & & \ding{51} & \ding{51} & \\
\bottomrule
\end{tabular*}
\end{NoHyper}
\end{table*}

\begin{table*}[t]
\centering
\caption{Task-specific success criteria for real-robot evaluation.
Success is determined based on the final physical state of the environment.}
\label{tab:success_criteria}

\small
\renewcommand{\arraystretch}{1.25}
\setlength{\tabcolsep}{4pt}

\begin{tabularx}{\linewidth}{@{}lX@{}}
\toprule
\textbf{Task} & \textbf{Success Criterion} \\
\midrule

Block Stacking &
The green block stays in place. Once released by the gripper, the yellow block rests stably on top of the green block without touching the tabletop. \\

\addlinespace[3pt]

Plug Insertion &
The pins on the plug enter the matching holes in the socket. \\

\addlinespace[3pt]

Hanoi Tower &
The peg passes completely through the hole in the block. \\

\addlinespace[3pt]

Block in Drawer &
The block is fully contained within the open drawer and rests flat on its floor. \\

\bottomrule
\end{tabularx}
\end{table*}

\Cref*{tab:prior_simulation_methods} compares real-to-sim methods for robotic manipulation. Agentic Real2Sim~\cite{chen2026agentic} reconstructs recorded interactions, Simify~\cite{kapelyukh2026test} searches for feasible object arrangements, and Re³Sim~\cite{han2026re3simgeneratinghighfidelitysimulation} generates demonstrations. SyncTwin~\cite{huang2026synctwinfastdigitaltwin}, and GaussTwin~\cite{cai2026gausstwin} synchronize simulated object states with the physical scene. Real-is-Sim~\cite{abouchakra2025realissimbridgingsimtorealgap} fits scene-specific Gaussians and provides virtual views and object poses from its synchronized twin for PushT. Among the approaches compared here, SpatialHarness is the first to use an online simulator to provide a vision-language model with spatial information on demand during real-world manipulation. SpatialHarness builds the scene from CAD models and maintains online synchronization without scene-specific training. The VLM can request virtual views that supplement physical camera observations, query object geometry, and define query points at arbitrary locations in an object’s local coordinate frame. SpatialHarness also flags potentially unreliable simulation outputs so the VLM can discount them before acting. We evaluate this interface on four real-robot tasks, from block stacking and plug insertion to fitting a holed block over a peg and opening a drawer to place a block inside.

We reproduced GaussTwin~\cite{cai2026gausstwin} on the plug insertion sequence in \Cref*{fig:gausstwin_plug_insertion}. Using an initial synchronized pair of \mbox{RGB-D} images from the external and wrist cameras, we fitted separate 3D Gaussian models~\cite{kerbl20233dgaussiansplattingrealtime} for the plug and socket, optimizing each for 2,200 iterations. Both GaussTwin and our method use the same pair of RGB-D images from the external and wrist cameras to align the plug and socket models with the physical scene at initialization. The tracker corrected both objects' pose estimates in each subsequent frame.

At corresponding time steps, we rendered both methods' scenes from these viewpoints and compared the renderings with the real images, enlarging the plug--socket region for display. In both views, the positions of the plug and socket in SpatialHarness remain consistent with their positions in the real images throughout the sequence. The rendered objects also have clear shapes and details.

\section{Real-World Experiments Details}

\subsection{Task Success Criteria}
We evaluate task success based on the final physical state of the real-world environment. As summarized in ~\Cref{tab:success_criteria}, each task has a predefined binary success criterion specifying the physical conditions. A trial is considered successful only when all task-specific conditions are satisfied. Otherwise, including cases of partial completion,
it is counted as a failure. The same success criteria are applied consistently across all experiments.

\subsection{Robot Action Protocol}
\label{sec:supp_action_interfaces}
The control frameworks used in our real-robot experiments
use different action interfaces. We summarize their commands
in Table~\ref{tab:control_commands} and describe their
execution below. For GPT-6.1 Sol high~\cite{openai2026gpt61sol} and GPT-6 Astra low~\cite{gpt6astra}, we adapt RoboCurve's GPT control pipeline~\cite{inspect-robots} to the low-level controller of our Franka Research 3 arm. Instead of specifying only a final target pose, each model outputs a sequence of translation and rotation increments for the intermediate steps of a motion. These increments are executed sequentially, enabling finer control and explicit planning of intermediate trajectories to help avoid collisions. However, in our real-robot tests, Gemini-Argon 4~\cite{kavukcuoglu2026gemini4argon} could not reliably generate a sequence of translation and rotation increments for the intermediate steps of a motion. We therefore
retain RoboCurve's Gemini control pipeline~\cite{inspect-robots}. The model
specifies a final target pose through \texttt{move\_to},
and intermediate poses are generated by numerical
interpolation using fixed geometric formulas. For agentic control, we integrate SpatialHarness into Harness VLA~\cite{zhang2026harnessvla} without modifying its action protocol. The planner continues to use the original  manipulation primitives to execute the planned actions.

\begin{table*}[t]
\centering
\caption{Action commands used by different control pipelines.}
\label{tab:control_commands}

\small
\setlength{\tabcolsep}{4pt}
\renewcommand{\arraystretch}{1.1}

\newcommand{\controllogo}[1]{%
    \raisebox{-0.25ex}{%
        \includegraphics[height=1.75ex]{#1}%
    }%
    \hspace{0.4em}%
}

\newcommand{\controlmethod}[2]{%
    \begin{tabular}[t]{@{}l@{}}
        \controllogo{#1}#2 \\
        \hspace*{1.6em}+ \textbf{SpatialHarness}%
    \end{tabular}%
}

\begin{NoHyper}
\begin{tabularx}{\textwidth}{
    @{}
    >{\raggedright\arraybackslash}p{0.235\textwidth}
    >{\hsize=0.98\hsize\raggedright\arraybackslash}X
    >{\hsize=0.98\hsize\raggedright\arraybackslash}X
    >{\hsize=1.04\hsize\raggedright\arraybackslash}X
    @{}
}
\specialrule{\heavyrulewidth}{0pt}{0pt}

\rule[-1.25ex]{0pt}{3.65ex}\textbf{Method} &
\textbf{Position Command} &
\textbf{Orientation Command} &
\textbf{Gripper Command} \\

\specialrule{\lightrulewidth}{0pt}{0pt}

\rowcolor{WAMGreen}
\multicolumn{4}{c}{%
    \rule[-1.25ex]{0pt}{3.65ex}%
    \textbf{\textit{Direct Control}}%
} \\

\specialrule{\lightrulewidth}{0pt}{0pt}
\addlinespace[5pt]

\begin{tabular}[t]{@{}l@{}}
    \controllogo{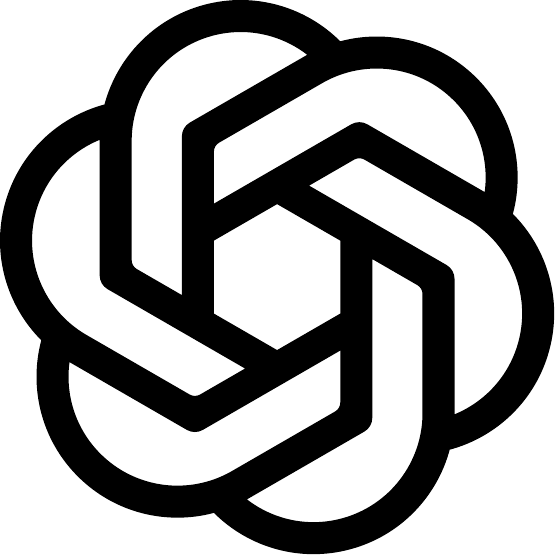}GPT-6.1 Sol high~\cite{openai2026gpt61sol} \\
    \controllogo{fig/appendix/gpt_logo.pdf}GPT-6 Astra low~\cite{gpt6astra} \\
    \hspace*{1.6em}+ \textbf{SpatialHarness}
\end{tabular} &
Apply a sequence of XYZ increments to specify intermediate positions during the motion. &
Apply a sequence of rotation-vector increments to specify intermediate orientations during the motion. &
Open or close the gripper with $+1$ or $-1$, respectively; use $0$ to maintain the current state. \\

\addlinespace[5pt]
\specialrule{\lightrulewidth}{0pt}{0pt}
\addlinespace[5pt]

\controlmethod{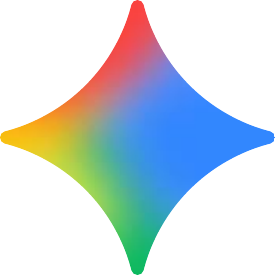}{Gemini-Argon 4 medium~\cite{kavukcuoglu2026gemini4argon}} &
Move the end effector to an absolute XYZ position via \texttt{move\_to} (m). &
Set the end effector to an absolute orientation via \texttt{move\_to} (rot6d). &
Set the gripper to open or closed with $1$ or $0$, respectively; omit the action to maintain the current state. \\

\addlinespace[5pt]
\specialrule{\lightrulewidth}{0pt}{0pt}

\rowcolor{VLABlue}
\multicolumn{4}{c}{%
    \rule[-1.25ex]{0pt}{3.65ex}%
    \textbf{\textit{Agentic Control}}%
} \\

\specialrule{\lightrulewidth}{0pt}{0pt}
\addlinespace[5pt]

\begin{tabular}[t]{@{}l@{}}
    \controllogo{fig/appendix/gpt_logo.pdf}Harness VLA~\cite{zhang2026harnessvla} \\
    \hspace*{1.6em}(GPT-6.1 Sol high~\cite{openai2026gpt61sol}) \\
    \hspace*{1.6em}+ \textbf{SpatialHarness}
\end{tabular} &
Move the end effector by a relative XYZ offset from its current position via \texttt{move\_delta} (m). &
Rotate the end effector relative to its current orientation via \texttt{rotate\_delta} (XYZ Euler angles, rad). &
Open or close the gripper via \texttt{open\_gripper} or \texttt{close\_gripper}, respectively; omit the action to maintain the current state. \\

\addlinespace[5pt]
\specialrule{\heavyrulewidth}{0pt}{0pt}
\end{tabularx}
\end{NoHyper}
\end{table*}

\begin{table*}[t]
  \centering
  \caption{Spatial Inspection Interfaces}
  \label{tab:planner-digital-twin-tools}
  \small
  \renewcommand{\arraystretch}{1.35}
  \begin{tabularx}{\linewidth}{@{}l@{\hspace{1em}}l@{\hspace{1.2em}}>{\raggedright\arraybackslash}X@{}}
    \toprule
    \textbf{Subcategory} & \textbf{Interface} & \textbf{Function} \\
    \midrule
    \rowcolor{black!7}[0pt][0pt]
    \multicolumn{3}{@{}l@{}}{\hspace{0.4em}\textbf{Virtual view rendering}} \\
    & \texttt{view\_scene}
      & Render the synchronized scene from a specified virtual camera pose. \\
    \addlinespace[2pt]
    \rowcolor{black!7}[0pt][0pt]
    \multicolumn{3}{@{}l@{}}{\hspace{0.4em}\textbf{Geometry query}} \\
    \hspace{0.8em}\textit{Scene inventory} & \texttt{list\_entities}
      & List the scene objects and their associated manipulation keypoints. \\
    \addlinespace[3pt]
    \hspace{0.8em}\textit{Entity state queries} & \texttt{get\_entity\_state}
      & Query an object's state (e.g., how far a drawer is open) or a keypoint's position. \\
    & \texttt{get\_object\_pose}
      & Retrieve an object's pose in the synchronized scene. \\
    & \texttt{get\_point\_position}
      & Retrieve the coordinates of a manipulation keypoint. \\
    \addlinespace[3pt]
    \hspace{0.8em}\textit{Geometric measurements} & \texttt{get\_object\_size}
      & Retrieve bounding-box information for an object. \\
    & \texttt{get\_point\_distance}
      & Measure the distance between two points. \\
    & \texttt{get\_angle}
      & Measure the angle between two directions to assess their alignment. \\
    \bottomrule
  \end{tabularx}
  \par\smallskip
  {\footnotesize\raggedright
   All tools query the latest scene estimate.\par
   Positions and lengths are
   expressed in meters, angles in radians, and orientations as quaternions
   in \([x, y, z, w]\) order.\par}
\end{table*}

\section{Analysis of Tool-Use Traces}
\begin{figure*}[p]
  \centering
  \makebox[\linewidth][c]{\includegraphics[width=1.16\linewidth]{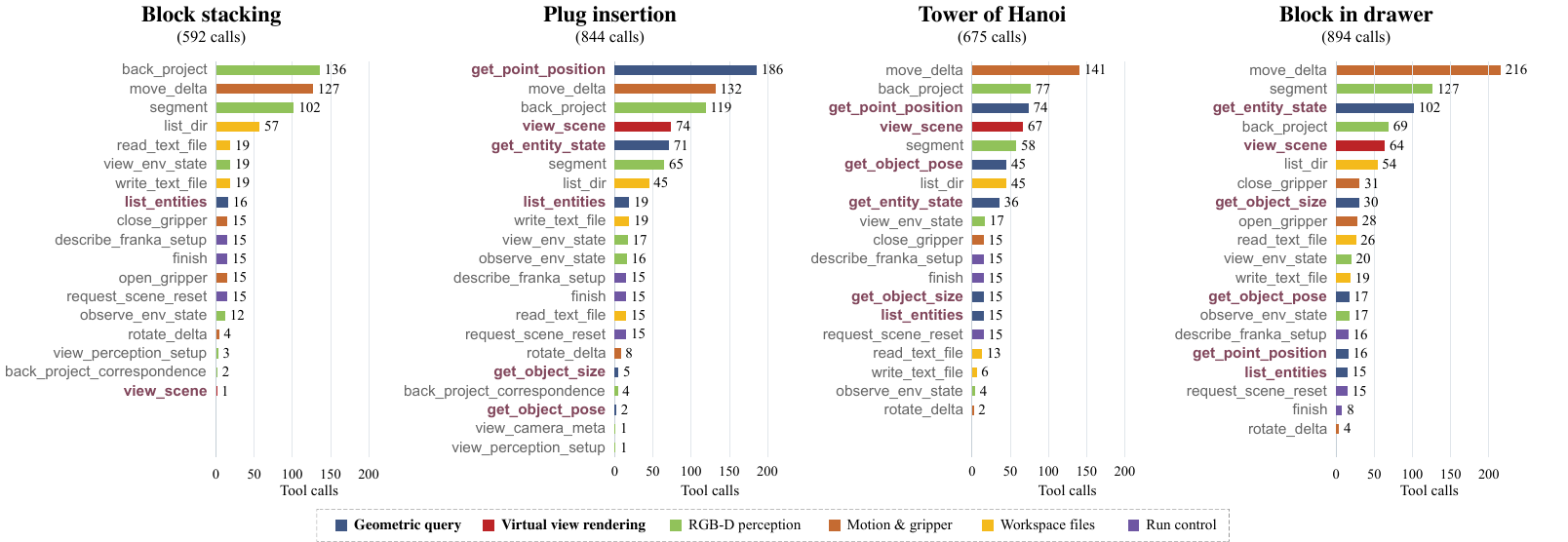}}
  \caption{Number of calls to each tool across the four tasks.}
  \label{fig:tool_calls_by_task}

  \vspace{26pt}
  \nextfloat
  \makebox[\linewidth][c]{\includegraphics[width=1.08\linewidth]{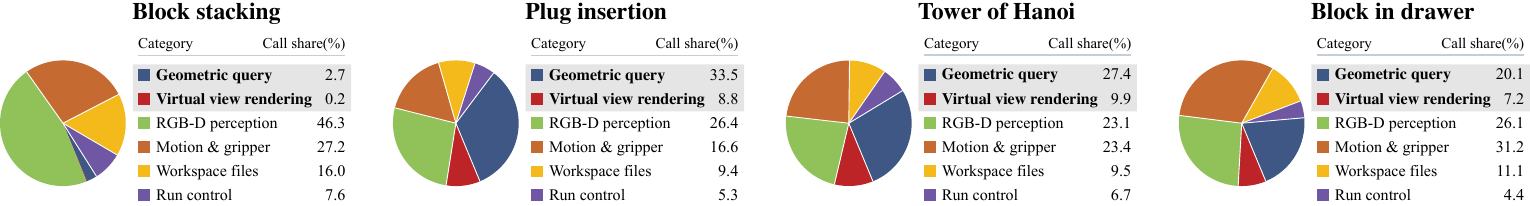}}
  \caption{Distribution of tool calls by category across the four tasks.}
  \label{fig:tool_call_distribution}
\end{figure*}

We visualize tool-use traces from our agentic-control experiments on the four tasks using Harness VLA augmented with SpatialHarness. \Cref{tab:planner-digital-twin-tools} lists the spatial inspection interfaces available to the planner. \Cref{fig:tool_calls_by_task} shows the number of calls to each tool across the four tasks, and \Cref{fig:tool_call_distribution} shows the distribution of tool calls by category. \Cref{fig:tool_trace_overall} summarizes transitions between consecutive tool calls across all tasks, while \Cref{fig:tool_trace_by_task} shows the corresponding transitions for each task separately.

For agentic control, geometric queries and virtual-view rendering together account for 2.9\%, 42.3\%, 37.3\%, and 27.3\% of all tool calls for Block Stacking, Plug Insertion, Hanoi Tower, and Block in Drawer, respectively. On the relatively simple stacking task, the agent makes little use of SpatialHarness. The baseline's 100\% success rate  with segmentation-based and depth-based perception primitives suggests that these basic tools provide sufficient information for this task and the finer spatial context from SpatialHarness is not necessary for this task completion. On the other three tasks, the agent uses a greater proportion of its tool calls to query geometry or request virtual views than on Block Stacking. With the additional spatial context from SpatialHarness, Harness VLA succeeds in a greater proportion of runs on all three tasks. Its success rates increase from 86.7\% to 100.0\% on Plug Insertion, from 60.0\% to 100.0\% on Hanoi Tower, and from 46.7\% to 86.7\% on Block in Drawer. In the aggregated plug-insertion traces, \texttt{get\_point\_position} is the most frequently called tool, with 186 calls. For Block in Drawer, \texttt{get\_entity\_state} is the most frequent geometric query, with 102 calls. These task-dependent patterns suggest that the agent can selectively retrieve spatial context suited to its information needs.

\section{Success and Failure Cases}
\begin{figure*}[p]
  \centering
  \includegraphics[width=0.92\linewidth]{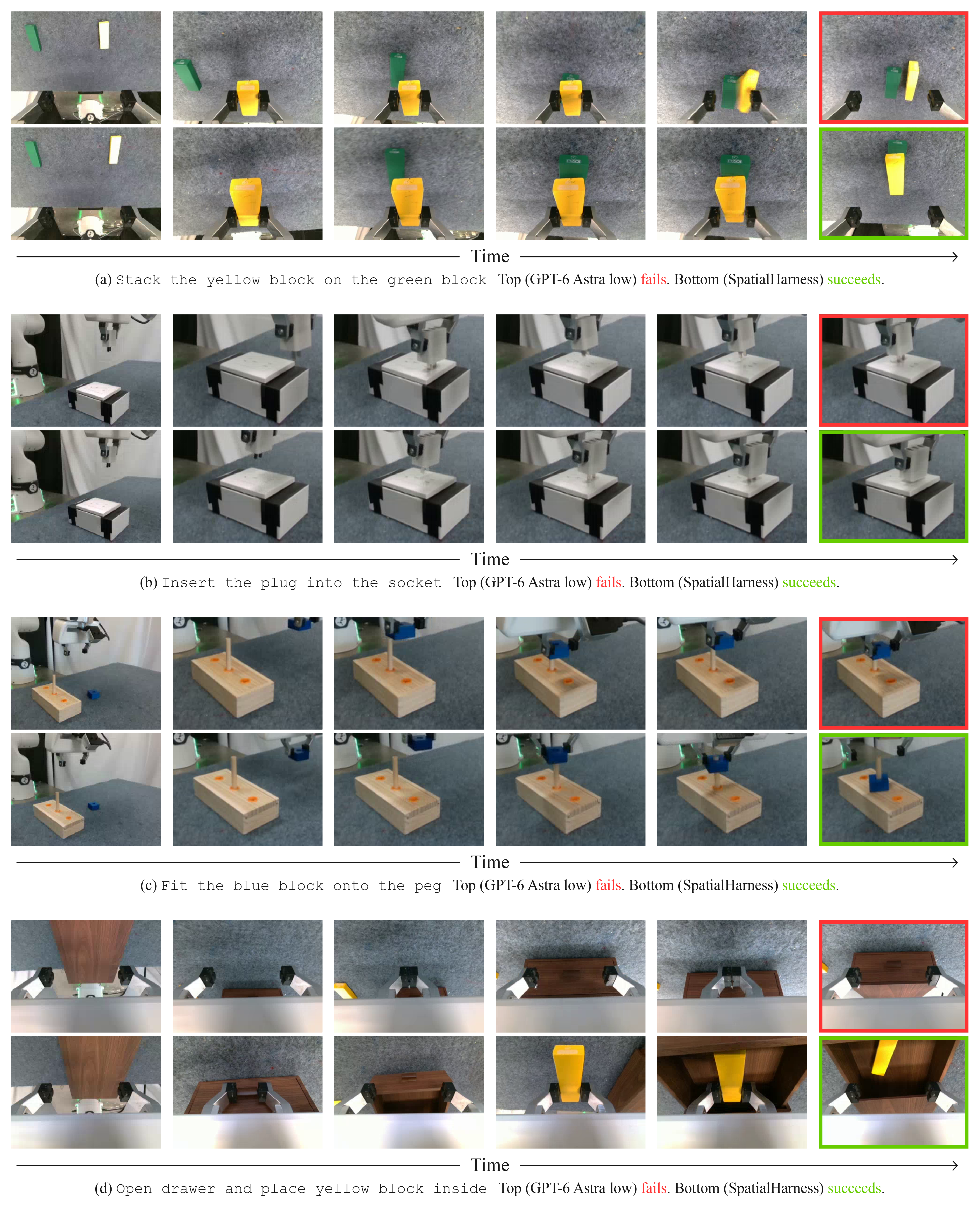}
  \caption{Representative rollouts under direct control using GPT-6 Astra with low reasoning effort. In the final column of each task, the red outline marks a \textcolor{FailureRed}{failed rollout} of baseline (top row) and the green outline marks a \textcolor{SuccessGreen}{successful rollout} of SpatialHarness (bottom row).
  From the second time point onward, panels~(b) and~(c) show zoomed-in views of both rollouts for easier visual comparison.}
  \label{fig:direct_rollouts}
\end{figure*}

\begin{figure*}[p]
  \centering
  \includegraphics[width=0.92\linewidth]{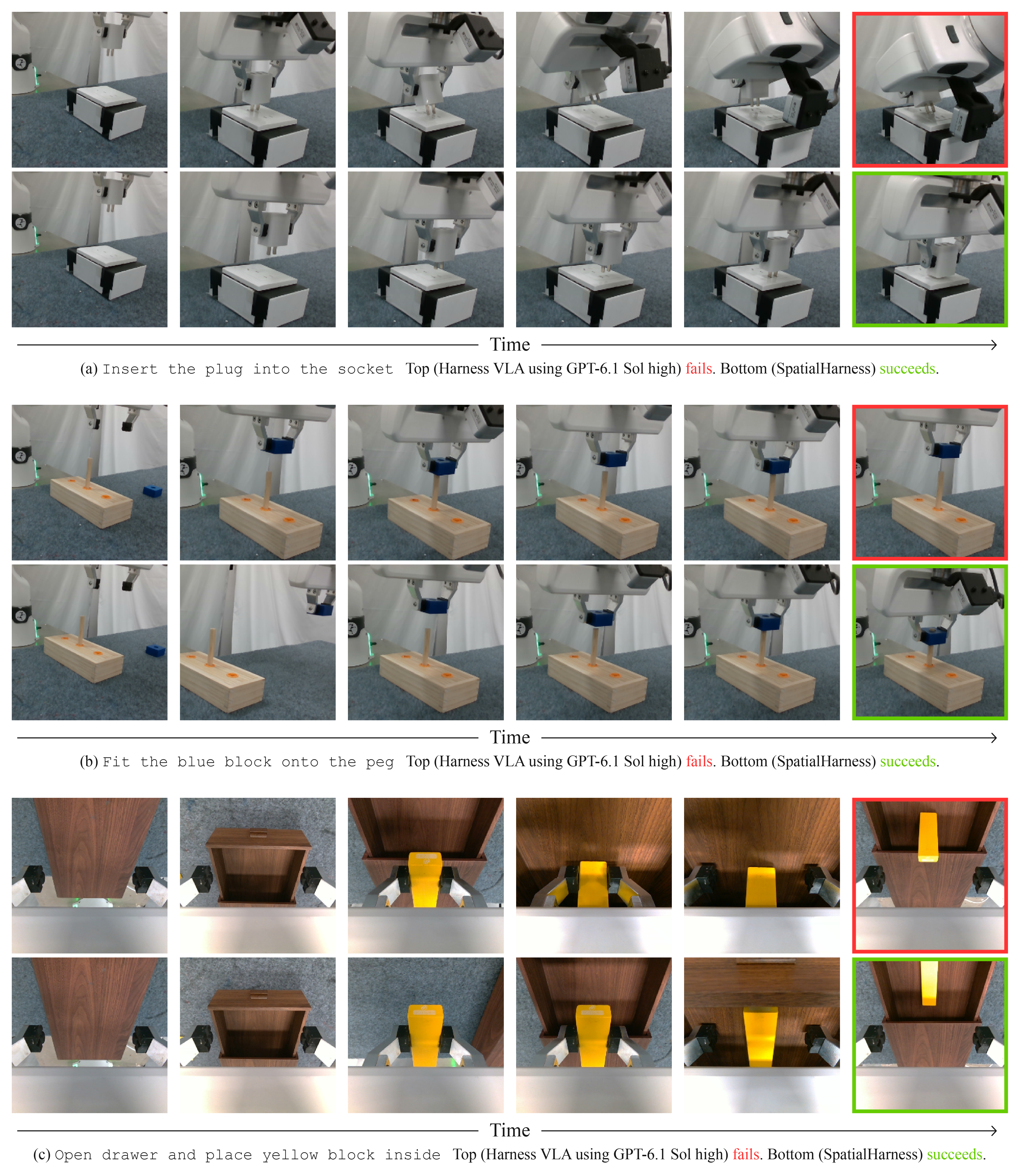}
  \caption{Representative rollouts under agentic control using Harness VLA with GPT-6.1 Sol at high reasoning effort as the action planner. In the final column of each task, the red outline marks a \textcolor{FailureRed}{failed rollout} of baseline (top row) and the green outline marks a \textcolor{SuccessGreen}{successful rollout} of SpatialHarness (bottom row).
  From the second time point onward, panels~(a) and~(b) show zoomed-in views of both rollouts for easier visual comparison.}
  \label{fig:agentic_rollouts}
\end{figure*}

\Cref{fig:direct_rollouts,fig:agentic_rollouts} compare representative failed baseline rollouts with successful SpatialHarness rollouts under direct and agentic control, respectively.

\noindent \textbf{Direct control.} In \Cref{fig:direct_rollouts}(a), the baseline places the yellow block in an unstable pose, causing it to tip off the green block when released. The other three baseline rollouts stall as the controller repeatedly tries to align the plug with the socket holes, fit the block over the peg, or grasp the drawer handle. With additional virtual views from SpatialHarness, the same controller completes all four tasks shown.

\noindent \textbf{Agentic control.} In the examples in \Cref{fig:agentic_rollouts}, alignment and clearance are difficult to judge from the physical cameras. The baseline continues insertion toward the wrong socket hole and collides with the socket, while the block strikes the drawer's front edge during placement. SpatialHarness provides additional views, such as a side view, and geometric measurements of object relationships, helping the agent avoid these collisions and complete the tasks.

\section{Visualizing Scene Synchronization Across Tasks}
\begin{figure*}[p]
  \centering
  \includegraphics[width=0.92\linewidth]{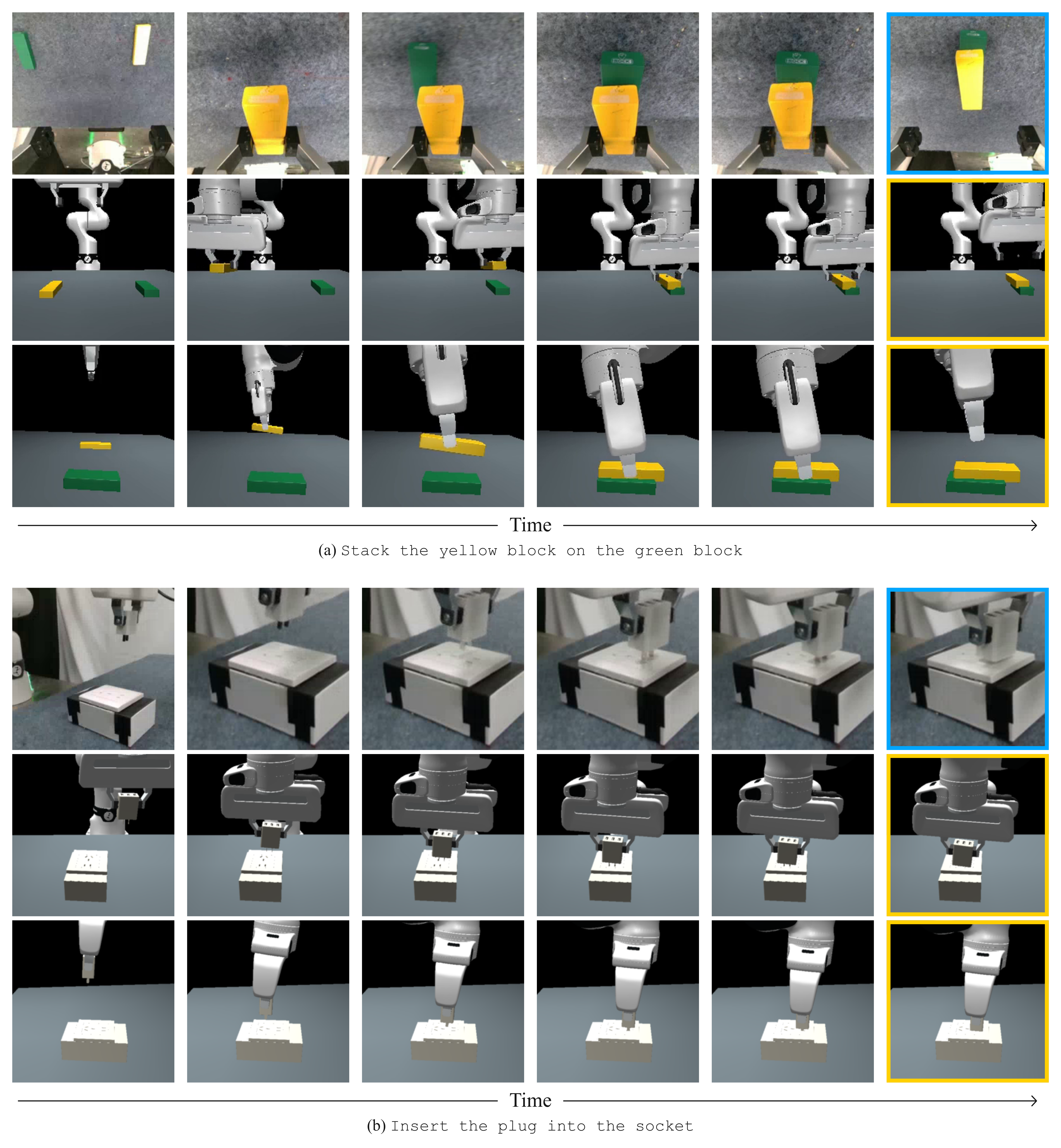}
  \caption{Scene synchronization for (a) block stacking and (b) plug insertion. Each row shows successive stages from left to right. In the final column of each task, the blue outline marks the \textcolor{ObservationBlue}{real observation} (top row) and the yellow outlines mark the two \textcolor{SimulationYellow}{simulated views} (lower rows).
  From the second time point onward, the top row of panel~(b) shows zoomed-in real observations for easier comparison with the simulated views.}
  \label{fig:scene_sync_tasks_1_2}
\end{figure*}

\begin{figure*}[p]
  \centering
  \includegraphics[width=0.92\linewidth]{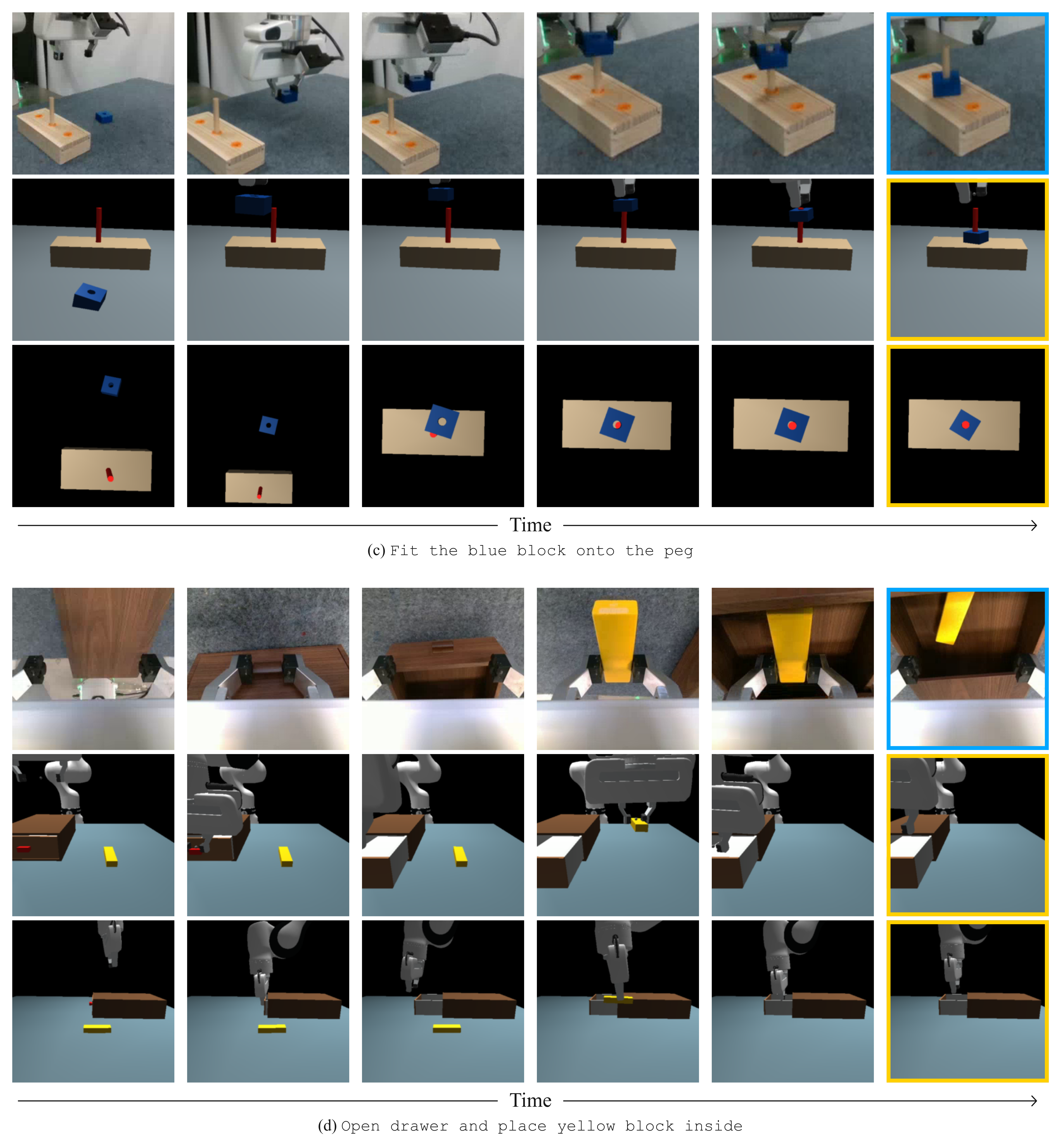}
  \caption{Scene synchronization for (c) Hanoi Tower and (d) block in drawer. Each row shows successive stages from left to right. In the final column of each task, the blue outline marks the \textcolor{ObservationBlue}{real observation} (top row) and the yellow outlines mark the two \textcolor{SimulationYellow}{simulated views} (lower rows).
  From the second time point onward, the top row of panel~(c) shows zoomed-in real observations for easier comparison with the simulated views.}
  \label{fig:scene_sync_tasks_3_4}
\end{figure*}

\Cref{fig:scene_sync_tasks_1_2,fig:scene_sync_tasks_3_4} show the physical execution (top row) and two virtual views of the synchronized scene (lower rows) for each task. Each row progresses from left to right.
The two virtual views are selected for each task before execution and rendered from the updated synchronized scene at each decision step for the direct control framework.

In Plug Insertion (\Cref{fig:scene_sync_tasks_1_2}(b)), the virtual side view lets the controller assess the plug's front-to-back alignment with the target socket holes and adjust it forward or backward before lowering it, helping avoid misaligned insertion.

In \Cref{fig:scene_sync_tasks_3_4}(c), the physical camera does not clearly reveal the peg tip's lateral position relative to the blue block's hole during placement. In the third column, the top-down view shows the tip outside the hole, guiding the controller to adjust the block horizontally. By the fourth column, the tip overlaps the hole, and the controller switches from horizontal adjustment to lowering the gripper vertically, placing the block over the peg. The side view confirms that the block reaches the base.

\clearpage

\begingroup
\makeatletter
\setlength{\@dblfpsep}{26pt}
\makeatother
\begin{figure*}[p]
  \centering
  \includegraphics[width=0.65\linewidth]{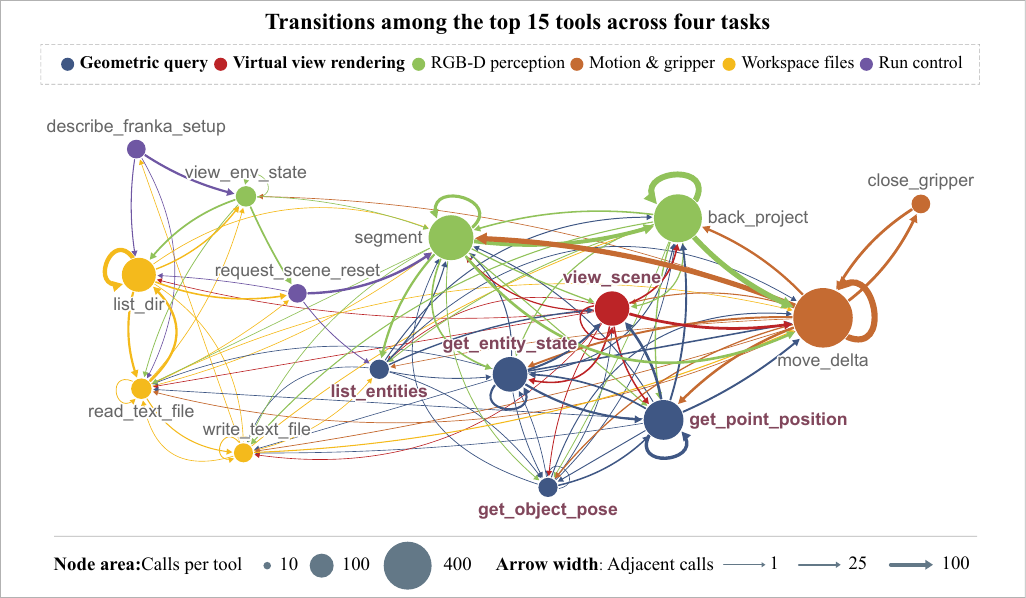}
  \caption{Aggregate transitions between consecutive tool calls across all four tasks. The graph shows transitions only when both tools are among the 15 most frequently called across the four tasks. Node area represents the number of calls to a tool, and arrow width represents the number of consecutive calls between tools.}
  \label{fig:tool_trace_overall}

  \vspace{26pt}
  \nextfloat
  \makebox[\linewidth][c]{\includegraphics[width=1.1\linewidth]{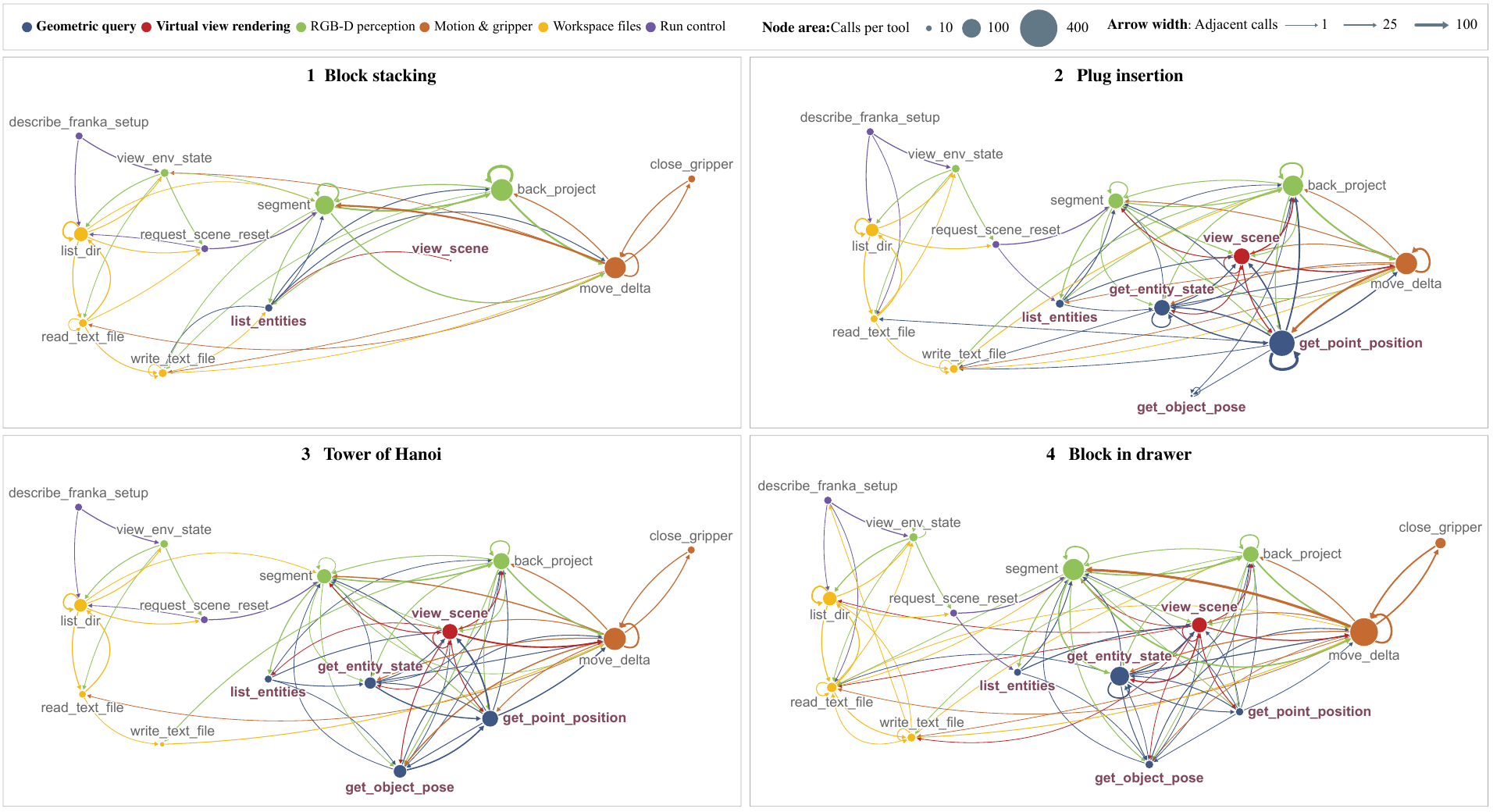}}
  \caption{Transitions between consecutive tool calls for each task. Each panel shows the tools used in that task among the 15 most frequently called across all four tasks. Node area represents the number of calls to a tool, and arrow width represents the number of consecutive calls between tools.}
  \label{fig:tool_trace_by_task}
\end{figure*}
\endgroup

\end{document}